\documentclass{article}
\pdfoutput=1
\usepackage{arxiv}

\usepackage[utf8]{inputenc} 
\usepackage[T1]{fontenc}    
\usepackage{hyperref}       
\usepackage{url}            
\usepackage{booktabs}       
\usepackage{amsfonts}       
\usepackage{amsmath}
\newcommand\numberthis{\addtocounter{equation}{1}\tag{\theequation}}

\usepackage{nicefrac}       
\usepackage{microtype}      
\usepackage{lipsum}
\usepackage{graphicx}
\usepackage{float}
\usepackage[natbibapa]{apacite}
\graphicspath{ {./images/} }

\usepackage{graphicx}
\usepackage[ruled,vlined,noend]{algorithm2e}
\usepackage{subcaption}
\usepackage{float}

\newcommand{\KL}{D_\mathrm{KL}}

\begin{document}

\title{Predictive Coding: a Theoretical and Experimental Review}
%
\author{
 Beren Millidge \\
  School of Informatics\\
  University of Edinburgh\\
  \texttt{beren@millidge.name} \\
  \And
  Anil K Seth \\
  Sackler Center for Consciousness Science\\
  School of Engineering and Informatics\\
 University of Sussex\\
 CIFAR Program on Brain, Mind, and Consciousness \\
\texttt{A.K.Seth@sussex.ac.uk} \\
  \And
    Christopher L Buckley \\
  Evolutionary and Adaptive Systems Research Group\\
  School of Engineering and Informatics\\
  University of Sussex\\
  \texttt{C.L.Buckley@sussex.ac.uk} 
  }
%
%
%
\maketitle   
%


\begin{abstract}

Predictive coding offers a potentially unifying account of cortical function -- postulating that the core function of the brain is to minimize prediction errors with respect to a generative model of the world. The theory is closely related to the Bayesian brain framework and, over the last two decades, has gained substantial influence in the fields of theoretical and cognitive neuroscience. A large body of research has arisen based on both empirically testing improved and extended theoretical and mathematical models of predictive coding, as well as in evaluating their potential biological plausibility for implementation in the brain and the concrete neurophysiological and psychological predictions made by the theory. Despite this enduring popularity, however, no comprehensive review of predictive coding theory, and especially of recent developments in this field, exists. Here, we provide a comprehensive review both of the core mathematical structure and logic of predictive coding, thus complementing recent tutorials in the literature \citep{buckley2017free, bogacz2017tutorial}. We also review a wide range of classic and recent work within the framework, ranging from the neurobiologically realistic microcircuits that could implement predictive coding, to the close relationship between predictive coding and the widely-used backpropagation of error algorithm, as well as surveying the close relationships between predictive coding and modern machine learning techniques.
\end{abstract}
\tableofcontents
\newpage
\section{Introduction}

Predictive coding theory is an influential theory in computational and cognitive neuroscience, which proposes a potential unifying theory of cortical function \citep{friston2003learning,friston2005theory,rao1999predictive,friston2010free,clark2013whatever,seth2014cybernetic} -- namely that the core function of the brain is simply to minimize prediction error, where the prediction errors signal mismatches between predicted input and the input actually received \footnote{For a contrary view and philosophical critique see \citet{cao2020new}.}. This minimization can be achieved in multiple ways: through immediate inference about the hidden states of the world, which can explain perception \citep{beal2003variational}, through updating a global world-model to make better predictions, which could explain learning \citep{friston2003learning,neal1998view}, and finally through action to sample sensory data from the world that conforms to the predictions \citep{friston2009reinforcement}, which potentially provides an account adaptive behaviour and control \citep{friston2015active}. Prediction error minimization can also be influenced by modulating the precision of sensory signals, which corresponds to modulating the `signal to noise ratio' in how prediction errors can be used to update prediction, and which may shed light on the neural implementation of attention mechanisms \citep{feldman2010attention,kanai2015cerebral}. Predictive coding boasts an extremely developed and principled mathematical framework in terms of a variational inference algorithm \citep{blei2017variational,ghahramani2000graphical, jordan1998introduction}, alongside many empirically tested computational models with close links to machine learning \citep{beal2003variational,dayan1995helmholtz,hinton1994autoencoders}, which address how predictive coding can be used to solve challenging perceptual inference and learning tasks similar to the brain. Moreover, predictive coding also has been translated into neurobiologically plausible microcircuit process theories \citep{bastos2012canonical,shipp2016neural,shipp2013reflections} which are increasingly supported by neurobiological evidence. Predictive coding as a theory also offers a single mechanism that accounts for diverse perceptual and neurobiological phenomena such as end-stopping \citep{rao1999predictive}, bistable perception \citep{hohwy2008predictive,weilnhammer2017predictive}, repetition suppression \citep{auksztulewicz2016repetition}, illusory motions \citep{lotter2016deep,watanabe2018illusory}, and attentional modulation of neural activity \citep{feldman2010attention,kanai2015cerebral}. As such, and perhaps uniquely among neuroscientific theories, predictive coding encompasses all three layers of Marr's hierarchy by providing a well-characterised and empirically supported view of `what the brain is doing' at all of the computational, algorithmic, and implementational levels \citep{marr1982vision}.

The core intuition behind predictive coding is that the brain is composed of a hierarchy of layers, which each make predictions about the activity of the layer immediately below them in the hierarchy \citep{clark2015surfing} \footnote{For much of this work we consider a simple hierarchy with only a single layer above and below. Of course, connectivity in the brain is more heterarchical with many `skip connections'. Predictive coding can straightforwardly handle these more complex architectures in theory, although few works have investigated the performance characteristics of such heterarchical architectures in practice.}. These downward descending predictions at each level are compared with the activity and inputs of each layer to form prediction errors -- which is the information in each layer which could not be successfully predicted. These prediction errors are then fed upwards to serve as inputs to higher levels, which can then be utilized to reduce their own prediction error. The idea is that, over time, the hierarchy of layers instantiates a range of predictions at multiple scales, from the fine details in local variations of sensory data at low levels, to global invariant properties of the causes of sensory data (e.g., objects, scenes) at higher or deeper levels\footnote{This pattern is widely seen in the brain \citep{hubel1962receptive,grill2004human} and also in deep (convolutional) neural networks \citep{olah2017feature}, but it is unclear whether this pattern also holds for deep predictive coding networks, primarily due to the relatively few instances of deep convolutional predictive coding networks in the literature so far.}. Predictive coding theory claims that goal of the brain as a whole, in some sense, is to minimize these prediction errors, and in the process of doing so performs both perceptual inference and learning. Both of these processes can be operationalized via the minimization of prediction error, first through the optimization of neuronal firing rates on a fast timescale, and then the optimization of synaptic weights on a slow timescale \citep{friston2008hierarchical}. Predictive coding proposes that using a simple unsupervised loss function, such as simply attempting to predict incoming sensory data, is sufficient to develop complex, general, and hierarchically rich representations of the world in the brain, an argument which has found recent support in the impressive successes of modern machine learning models trained on unsupervised predictive or autoregressive objectives \citep{radford2019language,kaplan2020scaling,brown2020language}. Moreover, in contrast to modern machine learning algorithms which are trained to end with a global loss at the output, in predictive coding prediction errors are computed at every layer which means that each layer only has to focus on minimizing local errors rather than a global loss. This property potentially enables predictive coding to learn in a biologically plausible way using only local and Hebbian learning rules \citep{whittington2017approximation,millidge2020predictive,friston2003learning}.

While predictive coding as a neuroscientific theory originated in the 1980s and 1990s \citep{srinivasan1982predictive,mumford1992computational,rao1999predictive}, and was first developed into its modern mathematical form of a comprehensive theory of cortical responses in the mid 2000s \citep{friston2003learning,friston2005theory,friston2008hierarchical}, it has deep intellectual antecedents. These precursors include Helmholtz's notion of perception as unconscious inference and Kant's notion that a priori structure is needed to make sense of sensory data \citep{hohwy2008predictive,seth2020preface}, as well as early ideas of compression and feedback control in cybernetics and information theory \citep{wiener2019cybernetics,shannon1948mathematical,conant1970every}.
One of the core notions in predictive coding is the idea that the brain encodes a model of the world (or more precisely, of the causes of sensory signals), which is used to make constant predictions about the world, which are then compared against sensory data. On this view, perception is not the result of an unbiased feedforward, or bottom-up, processing of sensory data, but is instead a process of using sensory data to update predictions generated internally by the brain. Perception, thus, becomes a `controlled hallucination' \citep{clark2013whatever,seth2020preface} in which top-down perceptual predictions are reined in by sensory prediction error signals. This view of `perception as unconscious inference' originated with the German physicist and physiologist Hermann von Helmholtz \citep{helmholtz1866concerning}, who studied the way the brain ``cancels out'' visual distortions and flow resulting from its own (predictable movement), such as during voluntary eye movements, but does not do so for external perturbations, such as when  external pressure is applied to the eyeball, in which case we consciously experience visual movement arising from this (unpredicted) ocular motion. Helmholtz thus argued that the brain must maintain both a record of its own actions, in the form of a `corollary discharge' as well as a model of the world sufficient to predict the visual effects of these actions (i.e. a forward model) in order to so perfectly cancel self-caused visual motion \citep{huang2011predictive}.

Another deep intellectual influence in predictive coding comes from information theory \citep{shannon1948mathematical}, and especially the minimum redundancy principle of Barlow \citep{barlow1961coding,barlow1961possible,barlow1989unsupervised}. Information theory tells us that information is inseparable from a lack of predictability. If something is predictable before observing it, it cannot give us much information. Conversely, to maximize the rate of information transfer, the message must be minimally predictable and hence minimally redundant. Predictive coding as a means to remove redundancy in a signal was first applied in signal processing, where it was used to reduce transmission bandwidth for video transmission. For a review see \citet{spratling2017review}. Initial schemes used a simple approach of subtracting the new (to-be-transmitted) frame from the old frame (in effect using a trivial prediction that the new frame is always the same as the old frame), which works well in reducing bandwidth in many settings where there are only a few objects moving in the video against a static background. More advanced methods often predict each new frame using a number of past frames weighted by some coefficient, an approach known as linear predictive coding. Then, as long as the coefficients are transmitted at the beginning of the message, the receiving system can reconstruct signals compressed by this system.
Barlow applied this principle to signalling in neural circuits, arguing that the brain faces considerably evolutionary pressures for information-theoretic efficiency, since neurons are energetically costly, and thus redundant firing would be potentially wasteful and damaging to an organism's evolutionary fitness. Because of this, we should expect the brain to utilize a highly optimized code which is minimally redundant. Predictive coding, as we shall see, precisely minimizes this redundancy, by only transmitting the errors or residuals of sensory input which cannot be explained by top-down predictions, thus removing the most redundancy possible at each layer \citep{huang2011predictive}. Finally, predictive coding also inherits intellectually from ideas in cybernetics, control and filtering theory \citep{kalman1960new,wiener2019cybernetics,conant1970every,seth2014cybernetic}. Cybernetics as a science is focused on understanding the dynamics of interacting feedback loops for perception and control, based especially around the concept of error minimization. Control and filtering theory have, in a related but distinct way, been based around methods to minimize residual errors in both perception or action according to some objective for decades. As we shall see, standard methods such as Kalman Filtering \citep{kalman1960new} or PID control \cite{johnson2005pid} can be shown as special cases of predictive coding under certain restrictive assumptions.

The first concrete discussion of predictive coding in the neural system arose as a model of neural properties of the retina \citep{srinivasan1982predictive}, specifically as a model of centre-surround cells which fire when presented with either a light-spot against a dark background (on-centre, off-surround), or alternatively a dark spot against a light background (off-centre, on surround) cells. It was argued that this coding scheme helps to minimize redundancy in the visual scene specifically by removing the \emph{spatial redundancy} in natural visual scenes -- that the intensity of one `pixel' helps predict quite well the intensities of neighbouring pixels. If, however, the intensity of a pixel was predicted by the intensity of the surround, and this prediction is subtracted from the actual intensity, then the centre-surround firing pattern emerges \citep{huang2011predictive}. Mathematically, this idea of retinal cells removing the spatial redundancy of the visual input is derived from the fact that the optimal spatial linear filter which minimizes the redundancy in the representation of the visual information closely resembles the centre-surround receptive fields which are well established in retinal ganglion cells \citep{huang2011predictive}. This predictive coding approach was also applied to coding in the lateral geniculate nucleus (LGN), the thalamic structure that retinal signals pass through en-route to cortex, which was hypothesised to help remove temporal correlations in the input by subtracting out the retinal signal at previous timesteps using recurrent lateral inhibitory connectivity \citep{huang2011predictive,marino2020predictive} 

\citet{mumford1992computational} was perhaps the first to extend this theory of the retina and the LGN to a fully-fledged general theory of cortical function. His theory was motivated by simple observations about the neurophysiology of cortico-cortical connections.  Specifically, the existence of separate feedforward and feedback paths, where the feedforward paths originated in the superficial layers of the cortex, and the feedback pathways originated primarily in the deep layers. He also noted the reciprocal connectivity observed almost uniformly between cortical regions -- if a region projects feedforward to another region, it almost always also receives feedback inputs from that region. He proposed that the deep feedback projections convey abstract `templates' which each cortical region then matches to its incoming sensory data. Then, inspired by the minimum redundancy principle of Barlow \citep{barlow1961possible}, he proposed that instead of faithfully transmitting the sensory input upwards, each layer transmits only the `residual' resulting after attempting to find the best fit match to the `template'. 

While Mumford's theory contained most aspects of classical predictive coding theory in the cortex, it was not accompanied by any simulations or empirical work and so its potential as a framework for understanding the cortex was not fully appreciated. The seminal work of Rao and Ballard in 1999 \citep{rao1999predictive} had its impact precisely by doing this. They created a small predictive coding network according to the principles identified by Mumford, and empirically investigated its behaviour, demonstrating that the complex and dynamic interplay of predictions and prediction errors could explain several otherwise perplexing neurophysiological phenomena, specifically `extra-classical' receptive field effects such as endstopping neurons. Extra-classical refers to the classical view in visual neuroscience of the visual system being composed of a hierarchy of feature-detectors, which originated in the pioneering work of \citep{hubel1962receptive}. According to this classical view, the visual cortex forms a hierarchy which ultimately bottoms out at the retina. At each layer, there are neurons sensitive to different features in the visual input, with neurons at the bottom of the hierarchy responding to simple features such as patches of light and dark, while neurons at the top respond to complex features such as faces. The feature detectors at higher levels of the hierarchy are computed by combining several lower-level simpler feature detectors. For instance, as a crude illustration, a face detector might be created by combining several spot detectors (eyes) with some bar detectors (mouths and noses). However, it was quickly noticed that some receptive fields displayed properties which could not be explained simply as compositions of lower-level feature detectors. Most significantly, many receptive field properties, especially in the cortex, showed context sensitivities, with their activity depending on the context outside of their receptive field. For instance, the `end-stopping' neurons fired if a bar was presented which ended just outside the receptive field of the cell, but not if it continued for a long distance beyond it. Within the classical feedforward view, such a feature detector should be impossible, since it would have no access to information outside of its receptive field. Rao and Ballard showed that a predictive coding network, constructed with both bottom up prediction error neurons and neurons providing \emph{top-down predictions}, enables the replication of several extra-classical receptive field properties, such as endstopping, within the network. This capability is made possible by the top-down predictions conveyed by the hierarchical predictive coding network. In effect, the predictive coding network conveys a downward prediction of the continuation of the bar, in line with ideas in gestalt perception. When this prediction is violated a prediction error is generated and the neuron fires, thus reproducing the extra-classical prediction error effect. Moreover, in the Rao and Ballard model prediction error, value estimation, and weight updates follow from gradient descents on a single energy function. This model was later extended by Karl Friston in a series of papers \citep{friston2003learning,friston2005theory,friston2008hierarchical}, which placed the model on a firm theoretical grounding as a variational inference algorithm, as well as integrating predictive coding with the broader free energy principle \citep{friston2006free,friston2010free} by identifying the energy function of Rao and Ballard with the variational free energy of variational inference. This identification enables us to understand the Rao and Ballard learning rules as performing well-specified approximate Bayesian inference.

Following the impetus of these landmark developments, as well as much subsequent work, predictive coding has become increasingly influential over the last two decades in cognitive and theoretical neuroscience, especially for its ability to offer a supposedly unifying, albeit abstract, perspective on the daunting multi-level complexity of the cortex. In this review, we aim to provide a coherent overview and introduction to the mathematical framework of predictive coding, as defined using the probabilistic modelling framework of \citet{friston2005theory}, as well as a comprehensive review of the many directions predictive coding theory has evolved in since then. For readers wishing to gain a deeper appreciation and understanding of the underlying mathematics, we also advise them to read these two didactic tutorials on the framework \citep{buckley2017free,bogacz2017tutorial}. We also advise reading \citet{spratling2017review} for a quick review of major predictive coding algorithms and \citet{marino2020predictive} for another overview of predictive coding and close investigation of its relationship to variational autoencoders \citep{kingma2013auto} and normalizing flows \citep{rezende2015variational}. In this review, we survey the development and performance of computational models designed to probe the performance of predictive coding on a wide variety of tasks, including those that try to combine predictive coding with ideas from machine learning to allow it to scale up to complex tasks. We also review the work that has been done on translating the relatively abstract mathematical formalism of predictive coding into hypothesized biologically plausible neural microcircuits that could, in principle, be implemented by the brain, as well as the empirical neuroscientific work explicitly seeking experimental confirmation or refutation of the many predictions made by the theory. We also look deeply at more theoretical matters, such as the extensions of predictive coding using dynamical models which utilizes generative models over multiple dynamical orders of motion, the relationship of learning in predictive coding to the backpropagation of error algorithm widely used in machine learning, and the development of the theory of precision which enables predictive coding to encode not just direct predictions of sensory stimuli but also predictions as to their intrinsic uncertainty. Finally, we review extensions of the predictive coding framework that generalize beyond perception to also include action, drawing on the close relationship between predictive coding and classical methods in filtering and control theory.

\section{Predictive Coding}

\subsection{Predictive Coding as Variational Inference}

A crucial advance in predictive coding theory occurred when it was recognized that the predictive coding algorithm could be cast as an approximate Bayesian inference process based upon Gaussian generative models \citep{friston2003learning,friston2005theory,friston2008hierarchical}. This perspective illuminates the close connection between predictive coding as motivated through the information-theoretic minimum-redundancy approach, and the Helmholtzian idea of perception as unconscious inference. Indeed, the two are fundamentally inseparable owing to the close mathematical connections between information theory and probability theory. Intuitively, information can only be defined according to some `expected' distribution, just as predictability or redundancy can only be defined against some kind of prediction. Prediction, moreover, presupposes some kind of model to do the predicting. The explicit characterisation of this model in probabilistic terms as a \emph{generative model} completes the link to probability theory and, ultimately Bayesian inference. Friston's approach, crucially, reformulates the mostly heuristic Rao and Ballard model in the language of variational Bayesian inference, thus allowing for a detailed theoretical understanding of the algorithm, as well as tying it the broader project of the Bayesian Brain \citep{deneve2005bayesian,knill2004bayesian}. Crucially, Friston showed that the energy function in Rao and Ballard can be understood as a variational free-energy of the kind that is minimized through variational inference. This connection demonstrates that predictive coding can be directly interpreted as performing approximate Bayesian inference to infer the causes of sensory signals, thus providing a mathematically precise characterisation of the Helmholtzian idea of perception as inference. 

Variational inference describes a broad family of methods which have been under extensive development in machine learning and statistics since the 1990s \citep{ghahramani2000graphical,jordan1998introduction,beal2003variational,blei2017variational}. They originally evolved out of methods for approximately solving intractable optimization problems in statistical physics \citep{feynman1998statistical}. In general, variational inference approximates an intractable inference problem with a tractable optimization problem. Intuitively, we postulate, and optimize the statistics of an approximate `variational' density, which we then try to match to the desired inference distribution \footnote{This contrasts with the other principle method for approximating intractable inference procedures -- Markov Chain Monte-Carlo (MCMC) \citep{brooks2011handbook,metropolis1953equation,hastings1970monte}. This class of methods sample stochastically from a Markov Chain with a stationary distribution equal to the true posterior. MCMC methods asymptotically converge to the true posterior, while variational methods typically do not (unless the class of variational distributions includes the true posterior). However, variational methods typically converge faster and are computationally cheaper, leading to a much wider use in contemporary machine learning and statistics.}.

To formalize this, let us assume we have some observations (or data) $o$, and we wish to infer the latent state $x$. We also assume we have a \emph{generative model} of the data generating process $p(o,x) = p(o | x)p(x)$. By Bayes rule, we can compute the true posterior directly as $p(x | o) = \frac{p(o,x)}{p(o)}$, however, the normalizing factor $p(o) = \int dx p(o,x)$ is often intractable because it requires an integration over all latent variable states. The marginal $p(o)$ is often referred to as the \emph{evidence}, since it effectively scores the likelihood of the data under a given model, averaged over all possible values of the model parameters. Computing the marginal $p(o)$ (\emph{model evidence}) is intrinsically valuable since it is a core quantity in Bayesian model comparison methods, where it is used to compare the ability of two different generative models to fit the data.


Since directly computing the true posterior $p(x | o)$ through Bayes rule is generally intractable, variational inference aims to approximate this posterior using an auxiliary posterior $q(x | o; \phi)$, with parameters $\phi$. This variational $q$ distribution is arbitrary and under the control of the modeller. For instance, suppose we define $q(x | o; \phi)$ to be a Gaussian distribution. Then the parameters $\phi=\{\mu, \Sigma\}$ become the mean $\mu$ and the variance $\Sigma$ of the Gaussian. The goal then is to fit this approximate posterior to the true posterior by minimizing the divergence between the true and approximate posterior with respect to the parameters. Mathematically, this problem can be written as, 
\begin{align*}
    q^*(x | o;\phi) = \underset{\phi}{argmin} \, \, \mathbb{D}[q(x | o; \phi) || p(x | o)] \numberthis
\end{align*}

Where $\mathbb{D}[p|q]$ is a function that measures the divergence between two distributions $p$ and $q$. Throughout, we take $\mathbb{D}[]$ to be the KL divergence $\mathbb{D}[Q ||P] = \KL[Q||P] = \int dQ Q \ln \frac{Q}{P}$, although other divergences are possible \citep{cichocki2010families,banerjee2005clustering} \footnote{Interestingly the KL divergence is asymmetric ($KL[Q||P] \neq KL[P||Q]$ and is thus not a valid distance metric. Throughout we use the reverse-KL divergence $KL[Q||P]$, as is standard in variational inference. Variational inference with the forward-KL $KL[P||Q]$ has close relationships to expectation propagation \citep{minka2001family}.}. When $\KL[q(x | o; \phi) || p(x | o)] = 0$ then $q(x | o; \phi) = p(x | o)$ and the variational distribution exactly equals the true posterior, and thus we have solved the inference problem\footnote{An exact solution is only possible when the family of variational distributions considered includes the true posterior as a member -- for example. if both the true posterior and the variational posterior are Gaussian.}. By doing this, we have replaced the inference problem of computing the posterior with an optimization problem of minimizing this divergence. However, merely writing the problem this way does not solve it because the divergence we need to optimize still contains the intractable true posterior. The beauty of variational inference is that it instead optimizes a tractable \emph{upper bound} on this divergence, called the \emph{variational free energy}\footnote{In machine learning, this is instead called the negative evidence lower bound (ELBO) which is simply the negative free-energy, and is maximized instead.}. To generate this bound, we simply apply Bayes rule to the true posterior to rewrite it in the form of a generative model and the evidence.
\begin{align*}
    \KL[q(x | o; \phi) || p(x | o)] &= \KL[q(x | o; \phi) || \frac{p(o,x)}{p(o)}] \\ 
    &= \KL[q(x | o; \phi) || p(o,x)] + \mathbb{E}_{q(x | o ;\phi)}[\ln p(o)] \\
    &= \KL[q(x | o; \phi) || p(o,x)] + \ln p(o) \\
    &\leq \KL[q(x | o; \phi) || p(o,x)] = \mathcal{F} \numberthis
\end{align*}

Where in the third line the expectation around $p(o)$ vanishes since the expectation is over the variable $x$ which is not in $p(o)$. The variational free energy $\mathcal{F} = \KL[q(x | o; \phi) || p(o,x)] $ is an upper bound because $\ln p(o)$ is necessarily $\leq 0$ since, as a probability, $0 \leq p(o) \leq 1$. Importantly, $\mathcal{F}$ \emph{is} a tractable quantity, since it is a divergence between two quantities we assume we (as the modeller) know -- the variational approximate posterior $q(x | o)$ and the generative model $p(o,x)$. Since $\mathcal{F}$ is an upper bound, by minimizing $\mathcal{F}$, we drive $q(x | o;\phi)$ closer to the true posterior. As an additional bonus, if $q(x | o; \phi) = p(x | o)$ then $\mathcal{F} = \ln p(o)$ or the marginal, or model, evidence, which means that in such cases $\mathcal{F}$ can  be used for model selection \citep{wasserman2000bayesian}. We can also gain an important intuition about $\mathcal{F}$ by showing that it can be decomposed into a likelihood maximization term and a KL divergence term which penalizes deviation from the Bayesian prior. These two terms are often called the `accuracy' and the `complexity' terms. This decomposition of $\mathcal{F}$ is often utilized and optimized explicitly in many machine learning algorithms \citep{kingma2013auto}.
\begin{align*}
    \mathcal{F} &= \KL[q(x | o; \phi) || p(o,x)] \\
    &= \underbrace{\mathbb{E}_{q(x | o; \phi)}\big[ \ln p(o | x) \big]}_{\text{Accuracy}} + \underbrace{\KL[q(x | o; \phi) || p(x)]}_{\text{Complexity}}
\end{align*}

In many practical cases, we must relax the assumption that we know the generative model $p(o,x)$. Luckily this is not fatal. Instead, it is possible to \emph{learn} the generative model alongside the variational posterior on the fly and in parallel using the Expectation Maximization (EM) algorithm \citep{dempster1977maximum}. The EM algorithm is extremely intuitive. First, assume that we parametrize our unknown generative model with some parameters $\theta$ which are initialized at some arbitrary $\theta_0$. Similarly, we initialize our variational posterior at some arbitrary $\phi_0$. Then, we take turns optimizing $\mathcal{F}$ with respect to the variational posterior parameters $\phi$ with the generative model parameters $\theta$ held fixed and then, conversely, optimize $\mathcal{F}$ with respect to the generative model parameters $\theta$ with the variational posterior parameters $\phi$ held fixed. Mathematically, we can write this alternating optimization as
\begin{align*}
    \phi_{t+1} &= \underset{\phi}{argmin} \, \mathcal{F}(\phi, \theta) \bigr|_{\theta = \theta_t} \\ 
    \theta_{t+1} &= \underset{\theta}{argmin} \, \mathcal{F}(\phi, \theta) \bigr|_{\phi = \phi_{t+1}} \numberthis
\end{align*}

Where we use the $\bigr|_{x=y}$ notation to mean that the variable $x$ is fixed at value $y$ throughout the optimization. It has been shown that this iterative sequence of optimization problems often converges to good solutions and often does so robustly and efficiently in practice \citep{dempster1977maximum,dellaert2002expectation,boyles1983convergence,gupta2011theory}.

Having reviewed the general principles of variational inference, we can see how they relate to predictive coding. First, to make any variational inference algorithm concrete, we must specify the forms of the variational posterior and the generative model. To obtain predictive coding, we assume a Gaussian form for the generative model $p(o,x ; \theta) = p(o | x;\theta)p(x;\theta) = \mathcal{N}(o; f(\theta_1 x), \Sigma_{1})\mathcal{N}(x; g(\theta_2 \bar{\mu}), \Sigma_2)$ where we first partition the generative model into likelihood $p(o|x;\theta)$ and prior $p(x;\theta)$ terms. The mean of the likelihood Gaussian distribution is assumed to be some function $f$ of the hidden states $x$, which can be parameterized with parameters $\theta$, while the mean of the prior Gaussian distribution is set to some arbitrary function $g$ of the prior mean $\bar{\mu}$. The variances of the two gaussian distributions of the generative model are denoted $\Sigma_1$ and $\Sigma_2$. We also assume that the variational posterior is a dirac-delta (or point mass) distribution $q(x | o;\phi) = \delta(x - \mu)$ with a center $\phi = \mu$\footnote{In previous works, predictive coding has typically been derived by assuming a Gaussian variational posterior under the Laplace approximation. This approximation effectively allows you to ignore the variance of the Gaussian and concentrate only on the mean. This procedure is essentially identical to the dirac-delta definition made here, and results in the same update scheme. However, the derivation using the Laplace approximation is much more involved so, for simplicity, here we use the Dirac delta definition. The original Laplace derivation can be found in Appendix A of this review --  see also \citet{buckley2017free} for a detailed walkthrough.}. 

Given these definitions of the variational posterior and the generative model, we can write down the concrete form of the variational free energy to be optimized. We first decompose the variational free energy into an `Energy' and an `Entropy' term
\begin{align*}
    \mathcal{F} &= \KL[q(x | o;\phi) || p(o,x ; \theta)] \\
    &= \underbrace{\mathbb{E}_{q(x | o;\phi)}[\ln q(x | o;\phi)]}_{\text{Entropy}} - \underbrace{\mathbb{E}_{q(x | o;\phi)}[\ln p(o,x ; \theta)]}_{\text{Energy}} \numberthis
\end{align*}

where, since the entropy of the dirac-delta distribution is 0 (it is a point mass distribution), we can ignore the entropy term and focus solely on writing out the energy.
\begin{align*}
    \underbrace{\mathbb{E}_{q(x | o;\phi)}[\ln p(o,x ; \theta)]}_{\text{Energy}} &= \mathbb{E}_{ \delta ( x - \mu)}[\ln \big( \mathcal{N}(o; f(\theta_1 x), \Sigma_{1})\mathcal{N}(x; g(\theta_2 \bar{\mu}), \Sigma_2)\big)] \\
    &= \ln \mathcal{N}(o; f(\mu, \theta_1), \Sigma_{1}) + \ln \mathcal{N}(\mu; g(\bar{\mu}, \theta_2), \Sigma_2) \\
    &= -\frac{(o-f(\mu, \theta_1))^2}{2\Sigma_1} - \frac{1}{2} \ln 2 \pi \Sigma_1  -\frac{(\mu - g(\bar{\mu}, \theta_2))^2}{2\Sigma_2} - \frac{1}{2} \ln 2 \pi \Sigma_2 \\
    &= -\frac{1}{2} \big[\Sigma_1^{-1} \epsilon_o^2 + \Sigma_2^{-1}\epsilon_x^2 + \ln 2\pi \Sigma_1 + \ln 2 \pi \Sigma_2\big] \numberthis
\end{align*}

where we define the `prediction errors' $\epsilon_o = o - f( \mu, \theta_1)$ and $\epsilon_x = \mu - g(\bar{\mu},\theta_2)$. We thus see that the energy term, and thus the variational free energy, is simply the sum of two squared prediction error terms, weighted by their inverse variances, plus some additional log variance terms. 

Finally, to derive the predictive coding update rules, we must make one additional assumption -- that the variational free energy is optimized using the method of gradient descent such that,
\begin{align*}
    \frac{d \mu}{dt}  = -\frac{\partial \mathcal{F}}{\partial \mu} \numberthis
\end{align*}
Given this, we can derive dynamics for all variables of interest ($\mu, \theta_1, \theta_2)$ by taking derivatives of the variational free energy $\mathcal{F}$. The update rules are as follows,
\begin{align*}
    \label{PC_update_rules}
    \frac{d\mu}{dt} &= -\frac{\partial \mathcal{F}}{\partial \mu} = \Sigma_1^{-1} \epsilon_o \frac{\partial f}{\partial \mu} \theta^T - \Sigma^{-1}_2 \epsilon_x \numberthis \\
    \frac{d\theta_1}{dt} &= \frac{\partial \mathcal{F}}{\partial \theta_1} = -\Sigma_1^{-1} \epsilon_o \frac{\partial f}{\partial \theta_1} \mu^T \numberthis \\
    \frac{d\theta_2}{dt} &= \frac{\partial \mathcal{F}}{\partial \theta_2} = -\Sigma_2^{-1} \epsilon_x \frac{\partial g}{\partial \theta_2} \bar{\mu}^T \numberthis
\end{align*}
Importantly, these update rules are very similar to the ones derived in \citet{rao1999predictive}, and therefore can be interpreted as recapitulating the core predictive coding update rules. Furthermore while it is possible to run the dynamics for the $\mu$ and the $\theta$ simultaneously, it is often better to treat predictive coding as an EM algorithm and alternate the updates. Empirically, it is typically best to run the optimization of the $\mu$s, with fixed $\theta$ until close to convergence, and then run the dynamics on the $\theta$ with fixed $\mu$ for a short while. This implicitly enforces a separation of timescales upon the model where the $\mu$ are seen as dynamical variables which change quickly while the $\theta$ are slowly-changing parameters. For instance, the $\mu$s are typically interpreted as rapidly changing neural firing rates, while the $\theta$s are the slowly changing synaptic weight values \citep{rao1999predictive,friston2005theory}.

Finally, we can think about how this derivation of predictive coding maps onto putative psychological processes of perception and learning. The updates of the $\mu$ can be interpreted as a process of perception, since the $\mu$ is meant to correspond to the estimate of the latent state of the environment generating the $o$ observations. By contrast, the dynamics of the $\theta$ can be thought of as corresponding to learning, since these $\theta$ effectively define the mapping between the latent state $\mu$ and the observations $o$.  Importantly, as will be discussed in depth later, these predictive coding update equations can be relatively straightforwardly mapped onto a potential network architecture which only utilizes local computation and plasticity -- thus potentially making it a good fit for implementation in the cortex.

\subsection{Multi-layer Predictive Coding}

The previous examples have only focused on predictive coding with a single level of latent variables $\mu_1$. However, the expressiveness of such a scheme is limited. The success of deep neural networks in machine learning have demonstrated that having hierarchical sets of latent variables is key to enabling methods to learn powerful abstractions and to handle intrinsically hierarchical dynamics of the sort humans intuitively perceive. The predictive coding schemes previously introduced can be straightforwardly extended to handle hierarchical dynamics of arbitrary depth, equivalently to deep neural networks in machine learning. This is done through postulating multiple layers of latent variables $x1 \dots x_L$ and then defining the generative model as follows,
\begin{align*}
    p(x_0 \dots x_L) = p(x_L)\prod_{l=0}^{L-1} p(x_{l} | x_{l+1}) \numberthis
\end{align*}
where $p(x_{l} | x_{l+1}) = \mathcal{N}(x_l; f_l(\theta_{l+1}, x_{l+1}, \Sigma_l)$ and the final layer $p(x_L) = \mathcal{N}(x_L | \bar{x_L}, \Sigma_L)$ has an arbitrary prior $\bar{x_L}$ and the latent variable at the bottom of the hierarchy is set to the observation actually received $x_0 = o$. Similarly, we define a separate variational posterior for each layer $q(x_{1:L} | o) = \prod_{l=1}^L \delta(x_l - \mu_l)$, then the variational free energy can be written as a sum of the prediction errors at each layer,
\begin{align*}
    \mathcal{F} = \sum_{l=1}^L \Sigma_l^{-1} \epsilon_l^2 + \ln 2\pi \text{det}(\Sigma_l) \numberthis
\end{align*}
where $\epsilon_l$ = $\mu_l - f_l(\theta_{l+1}, \mu_{l+1})$ and $\text{det}(\Sigma)$ denotes the determinant of the covariane matrix $\Sigma$. Given that the free energy divides nicely into the sum of layer-wise prediction errors, it comes as no surprise that the dynamics of the $\mu$ and the $\theta$ are similarly separable across layers.

\begin{align*}
    \frac{d\mu_l}{dt} &= -\frac{\partial \mathcal{F}}{\partial \mu_l} = \Sigma_{l-1}^{-1} \epsilon_{l-1} \frac{\partial f_{l-1}}{\partial \mu_l} \theta_{l}^T - \Sigma^{-1}_l \epsilon_l \numberthis \\
    \frac{d\theta_l}{dt} &= -\frac{\partial \mathcal{F}}{\partial \theta_l} = \Sigma_l^{-1} \epsilon_{l-1} \frac{\partial f_{l-1}}{\partial \theta_l} \mu_l \numberthis
\end{align*}

We see that the dynamics for the variational means $\mu$ depend only on the prediction errors at their layer and the prediction errors on the level below. Intuitively, we can think of the $\mu$s as trying to find a compromise between causing error by deviating from the prediction from the layer above, and adjusting their own prediction to resolve error at the layer below. In a neurally-implemented hierarchical predictive coding network, \emph{prediction errors} would be the only information transmitted `upwards' from sensory data towards latent representations, while predictions would be transmitted downwards. Crucially for conceptual readings of predictive coding, this means that sensory data is \emph{not} directly transmitted up through the hierarchy, as is assumed in much of perceptual neuroscience. The dynamics for the $\mu$s are also fairly biologically plausible as they are effectively just the sum of the precision-weighted prediction errors from the $\mu$s own layer and the layer below, the prediction errors from below being transmitted back upwards through the synaptic weights $\theta^T$ and weighted with the gradient of the activation function $f_l$. This means that there is no direct feedforward pass, as is often assumed in models of vision, in predictive coding. It is possible, however, to augment predictive coding models with a feedforward pass, as is discussed in the section on hybrid inference.

Importantly, the dynamics for the synaptic weights are entirely local, needing only the prediction error from the layer below and the current $\mu$ at the given layer. The dynamics thus becomes a Hebbian rule between the presynaptic $\epsilon_{l-1}$ and postsynaptic $\mu_l$, weighted by the gradient of the activation function.

\begin{figure}
    \centering
    \includegraphics[scale=0.2]{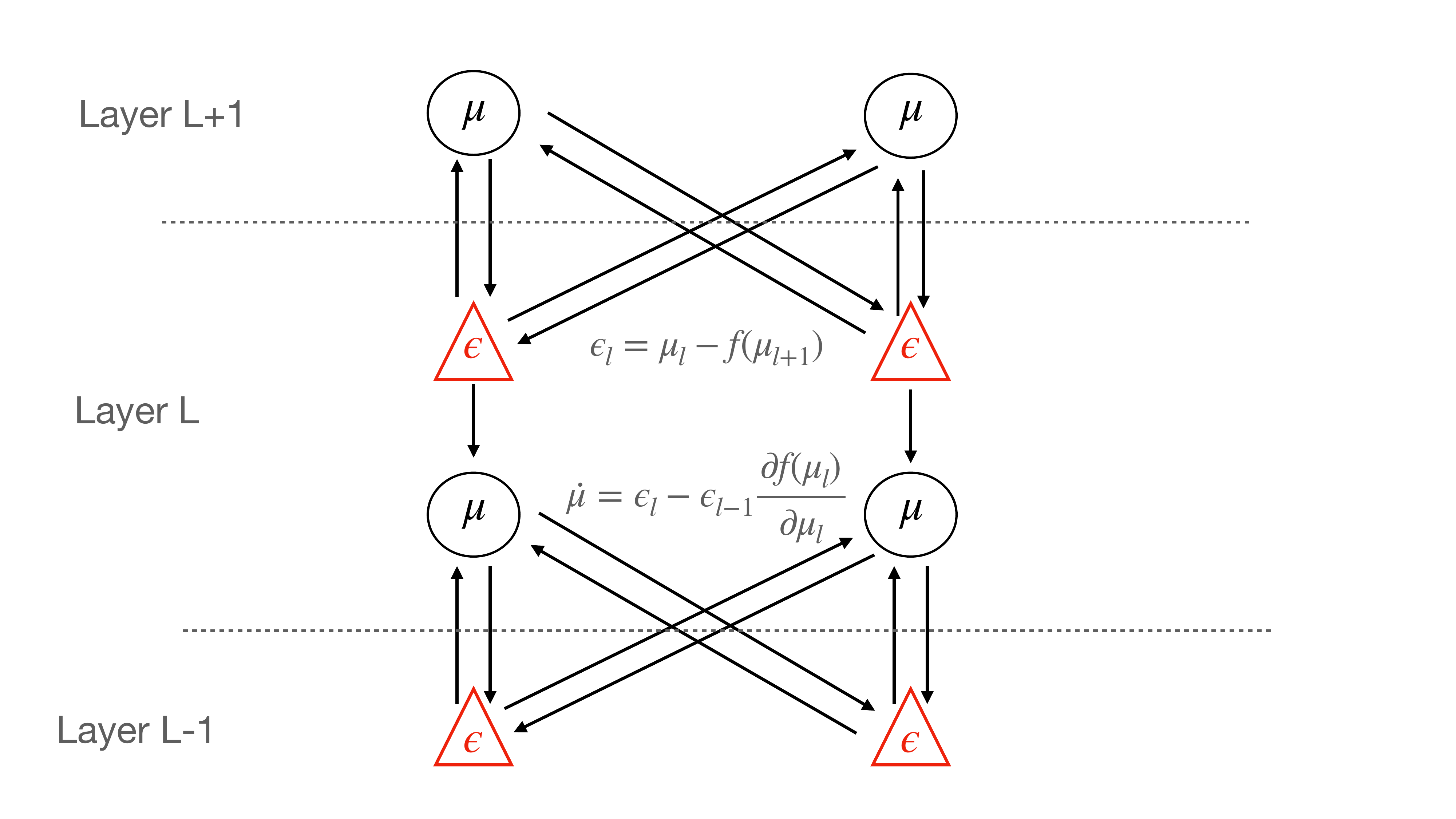}
    \caption{Architecture of a multi-layer predictive coding network (here shown with two value and error neurons in each layer. The value neurons $\mu$ project to both the error neurons of the layer below (representing the top down connections) and the error neurons at the current layer to represent the current activity. The error neurons receive inhibitory top down inputs from the value neurons of the layer above and excitatory inputs from the value neurons at the same layer. Conversely, the value neurons receive excitatory projections from the error neurons of the layer below and inhibitory from the error neurons at the current layer. Crucially, for this model with its explicit error neurons, all synaptic plasticity rules are purely Hebbian.}
    \label{fig:my_label}
\end{figure}

\subsection{Dynamical Predictive Coding and Generalized Coordinates} 

So far, we have considered the modelling of just a single static stimulus $o$. However, most interesting data the brain receives comes in temporal sequences $\bar{o} = [o_1, o_2 \dots ] $. To model such temporal sequences, it is often useful to split the latent variables into states, which can vary with time, and parameters which cannot. In the case of sequences, instead of minimizing the variational free energy, we must instead minimize the \emph{free action} $\bar{\mathcal{F}}$, which is simply the path integral of the variational free energy through time \footnote{This quantity is called the free action due to the analogy between it and the action central to the variational principles central to classical mechanics.} \citep{friston2008DEM,friston2008hierarchical}:
\begin{align*}
    \mu^* &= \underset{\mu}{argmin} \, \, \bar{\mathcal{F}}  \\
    \bar{\mathcal{F}}  &= \int dt \mathcal{F}_t \numberthis
\end{align*}

While there are numerous methods and parameterisations to handle sequence data, one influential and elegant approach, which has been developed by Friston in a number of key papers \citep{friston2008DEM,friston2008hierarchical,friston2010generalised} is to represent temporal data in terms of \emph{generalized coordinates of motion}. In effect, these represent not just the immediate observation state, but all time derivatives of the observation state. For instance, suppose that the brain represents beliefs about the position of an object. Under a generalized coordinate model, it would also represent beliefs about the velocity (first time derivative), acceleration (second time derivative), jerk (third time derivative) and so on. All these time derivative beliefs are concatenated to form a generalized state. The key insight into this dynamical formulation is, that when written in such a way, many of the mathematical difficulties in handling sequences disappear, leaving relatively straightforward and simple variational filtering algorithms which can natively handle smoothly changing sequences. 

Because generalised coordinates can become notationally awkward, we will be very explicit in the following. We denote the time derivatives of the generalized coordinate using a $'$, so $\mu'$ is the belief about the velocity of the $\mu$, just as $\mu$ is the belief about the `position' about the $\mu$. A key point of confusion is that there is also a `real' velocity of $\mu$, which we denote $\dot{\mu}$, which represents how the belief in $\mu$ actually changes over time -- i.e. over the course of inference. Importantly, this is not necessarily the same as the belief in the velocity: $\dot{\mu} \neq \mu'$, except at the equilibrium state. Intuitively, this makes sense as at equilibrium (minimum of the free action, and thus perfect inference), our belief about the velocity of mu $\mu'$ and the `real' velocity perfectly match. Away from equilibrium, our inference is not perfect so they do not necessarily match. We denote the generalized coordinate representation of a state $\tilde{\mu}$ as simply a vector of each of the beliefs about the time derivatives $\tilde{\mu} = [\mu, \mu', \mu'', \mu''' \dots]$. We also define the operator $\mathcal{D}$ which maps each element of the generalised coordinate to its time derivative i.e. $\mathcal{D}\mu = \mu', \mathcal{D}\tilde{\mu} = [\mu', \mu'', \mu''',\mu'''' \dots]$. With this notation, we can define a dynamical generative model using generalized coordinates. Crucially, we assume that the noise $\omega$ in the generative model is not white noise, but is colored, so it has non-zero autocorrelation and can be differentiated. Effectively, colored noise allows one to model relatively slowly (not infinitely fast) exogenous forces on the system. For more information on colored noise vs white noise see \citet{stengel1986stochastic,jazwinski2007stochastic,friston2008DEM}. With this assumption we can obtain a generative model in generalized coordinates of motion by simply differentiating the original model.
\begin{align*}
    o &= f(x) + \omega_o  && x = g(\bar{x}) + \omega_x \\
    o' &= f(x)x' + \omega_o' && x' = g(\bar{x})x' + \omega_x'\\
    o'' &= f(x)x'' + \omega_o'' && x'' = g(\bar{x})x'' + \omega_x'' \\
    &\dots && \dots \numberthis
\end{align*}
Where we have applied a local linearisation assumption \citep{friston2008DEM} which drops the cross terms in the derivatives. We can write these generative models more compactly in generalized coordinates.
\begin{align*}
    \tilde{o} = \tilde{f}(\tilde{x}) + \tilde{\omega}_o && \tilde{x} = \tilde{g}(\tilde{\bar{x}}) + \tilde{\omega}_x \numberthis
\end{align*}
which, written probabilistically is $p(\tilde{o},\tilde{x}) = p(\tilde{o} | \tilde{x})p(\tilde{x})$. It has been shown \citep{friston2008DEM} that the optimal (equilibrium) solution to this free action is the following stochastic differential equation,
\begin{align*}
    \label{DEM_SDE}
    \dot{\tilde{\mu}} = \mathcal{D}\tilde{\mu} + \frac{\partial \mathbb{E}_{q(\tilde{x} | \tilde{o}; \tilde{\mu})}[\ln p(\tilde{o}, \tilde{x})]}{\partial \tilde{\mu}} + \tilde{\omega} \numberthis
\end{align*}
Where $\tilde{\omega}$ is the generalized noise at all orders of motion. Intuitively, this is because when $\frac{\partial \mathbb{E}_{q(x | o; \mu)}[\ln p(\tilde{o}, \tilde{x})]}{\partial \mu} = 0$ then $\dot{\tilde{\mu}} = \mathcal{D}\tilde{\mu}$, or that the `real' change in the variable is precisely equal to the expected change. This equilibrium is a dynamical equilibrium which moves over time, but precisely in line with the beliefs $\mu'$. This allows the system to track a dynamically moving solution precisely, and the generalized coordinates let us capture this motion while retaining the static analytical approach of an equilibrium solution, which would otherwise necessarily preclude motion. There are multiple options to turn this result into a variational inference algorithm. Note, the above equation makes no assumptions about the form of variational density or the generative model, and thus allows multimodal or nonparametric distributions to be represented. For instance, the above equation Equation \ref{DEM_SDE} could be integrated numerically by a number of particles in parallel, thus leading to a generalization of particle filtering \citep{friston2008variational}. Alternatively, a fixed Gaussian form for the variational density can be assumed, using the Laplace approximation. In this case, we obtain a very similar algorithm to predictive coding as before, but using generalized coordinates of motion. In the latter case, we can write out the free energy as,
\begin{align*}
    \mathcal{F}_t &= \ln p(\tilde{o} | \tilde{x})p(\tilde{x}) \\
    &\propto \tilde{\Sigma}^{-1}_o \tilde{\epsilon}_o^2 + \tilde{\Sigma}^{-1}_x \tilde{\epsilon}_x^2  \numberthis
\end{align*}
Where $\tilde{\epsilon}_o =  \tilde{o} - \tilde{f}(\tilde{x})$ and $\tilde{\epsilon}_x =  \tilde{o} - \tilde{g}(\tilde{\bar{x}})$. Moreover, the generalized precisions $\tilde{\Sigma}^{-1}$ not only encode the covariance between individual elements of the data or latent space at each order, but also the correlations between generalized orders themselves. Since we are using a unimodal (Gaussian) approximation, instead of integrating the stochastic differential equations of multiple particles, we instead only need to integrate the deterministic differential equation of the mode of the free energy,
\begin{align*}
    \dot{\tilde{\mu}} = \mathcal{D}\tilde{\mu} - \tilde{\Sigma}^{-1}_o \tilde{\epsilon}_o - \tilde{\Sigma}^{-1}_x \tilde{\epsilon}_x \numberthis
\end{align*}
which cashes out in a scheme very similar to standard predictive coding (compare to Equation \ref{PC_update_rules}), but in generalized coordinates of motion. The only difference is the $\mathcal{D}\tilde{\mu}$ term which links the orders of motion together. This term can be intuitively understood as providing the `prior motion' while the prediction errors provide `the force' terms. To make this clearer, let's take a concrete physical analogy where $\mu$ is the position of some object and $\mu'$ is the expected velocity. Moreover, the object is subject to forces $\tilde{\Sigma}^{-1}_o \tilde{\epsilon}_o + \tilde{\Sigma}^{-1}_x \tilde{\epsilon}_x$ which instantaneously affect its position. Now, the total change in position $\dot{\tilde{\mu}}$ can be thought of as first taking the change in position due to the intrinsic velocity of the object $\mathcal{D}\mu$ and adding that on to the extrinsic changes due to the various exogenous forces. 

\subsection{Predictive Coding and Precision}
One core aspect of the predictive coding framework, which is absent in the original Rao and Ballard formulation, but which arises directly from the variational formulation of predictive coding and the Gaussian generative model, is the notion of \emph{precision} or inverse-variances, which we have throughout denoted as $\Sigma^{-1}$ (sometimes $\Pi$ is used in the literature as well). Precisions serve to multiplicatively modulate the importance of the prediction errors, and thus possess a significant influence in the overall dynamics of the model. They have been put to a wide range of theoretical purposes in the literature, all centered around their modulatory function. Early work \citep{friston2005theory} ties the precision parameters to lateral inhibition and biased competition models, proposing that they serve to mediate competition between prediction error neurons, and are implemented through lateral synaptic weights. Later work \citep{kanai2015cerebral,feldman2010attention} has argued instead that precisions can be interpreted as implementing top-down attentional modulation of predictions -- which are thus sensitive to the global context variables such as task relevance which have been shown empirically to have a large affect on attentional salience. This work has shown that equipping predictive coding schemes with precision allows them to recapitulate key phenomena observed in standard attentional psychophysics tasks such as the Posner paradigm \citep{feldman2010attention}.

Further theoretical and philosophical work has further developed the interpretation of precision matrices into a general purpose modulatory function \citep{clark2015surfing}. This function could be implemented neurally in a number of ways. First, certain precision weights could be effectively hardcoded by evolution. One promising candidate for this would be the precisions of interoceptive signals transmitting vital physiological information such as hunger or pain. These interoceptive signals would be hardwired to have extremely high precision, to prevent the organism from simply learning to ignore or down-weight them in comparison to other objectives. Conceptually, assuming high precision for interoceptive signals can shed light on how such signals drive adaptive action through active inference \citep{seth2016active,seth2013extending}. It is also possible that certain psychiatric disorders such as autism \citep{van2013predictive,lawson2014aberrant} and schizophrenia \citep{sterzer2018predictive} could be interpreted as disorders of  precision (either intrinsically hardcoded, or else resulting from an aberrant or biased learning of precisions). If on track, these theories would provide us with a helpful mechanistic understanding of the algorithmic deviations underlying these disorders, which could potentially enable improved differential diagnosis, and could even guide clinical intervention.  This approach, under the name of computational psychiatry, is an active area of research and perhaps one of the most promising avenues for translating highly theoretical models such as predictive coding into concrete medical advances and treatments \citep{huys2016computational}.

Mathematically, a dynamical update rule for the precisions can be derived as a gradient descent on the free-energy. This update rule becomes an additional M-step in the EM algorithm, since the precisions are technically parameters of the generative model. Recall that we can write the free-energy as,
\begin{align*}
    \mathcal{F} = \sum_{l=1}^L \Sigma_l^{-1} \epsilon_l^2 + \ln 2\pi \text{det}(\Sigma_l) \numberthis
\end{align*}

We can then derive the dynamics of the precision $\Sigma^{-1}_l$ matrix as a gradient descent on the free-energy with respect to the variance,
\begin{align*}
    \label{Precision_update}
    \frac{d \Sigma}{dt} = -\frac{\partial \mathcal{F}}{\partial \Sigma} &= \Sigma^{-T}_l \epsilon_l \epsilon_l^T \Sigma^{-T}_l - \Sigma_l^{-T} \\
    &= \Sigma^{-1}_l \epsilon_l \epsilon_l^T \Sigma^{-T}_l - \Sigma_l^{-1} \\
    &= \tilde{\epsilon_l} \tilde{\epsilon_l}^T - \Sigma_l^{-1} \numberthis
\end{align*}

Where we have used the fact that the covariance (and precision) matrices are necessarily symmetric ($\Sigma^{-1} = \Sigma^{-T}$). Secondly, we have defined the precision-weighted predictions errors as $\tilde{\epsilon}_l = \Sigma^{-1}_l \epsilon_l$. From these dynamics, we can see that the average fixed-point of the precision matrix is simply the variance of the precision-weighted prediction errors at each layer
\begin{align*}
    \mathbb{E}[\frac{d \Sigma}{dt}] &= \mathbb{E}[\tilde{\epsilon_l} \tilde{\epsilon_l}^T] - \Sigma_l \\
    &= \mathbb{E}[\frac{d \Sigma}{dt}] = 0 \implies \Sigma_l = \mathbb{E}[\tilde{\epsilon_l} \tilde{\epsilon_l}^T] 
    \\
    &\implies \Sigma_l = \mathbb{V}[\tilde{\epsilon}] \numberthis
\end{align*}

In effect, the fixed point of the precision dynamics will lead to these matrices simply representing the average variance of the prediction errors at each level. At the lowest level of the hierarchy, the variance of the prediction errors will be strongly related to the intrinsic variance of the data, and thus algorithmically, learnable precision matrices allow the representation and inference on data with state-dependent additive noise. This kind of state-dependent noise is omnipresent in the natural world, in part due to its own natural statistics, and in part due to  intrinsically noisy properties of biological perceptual systems \citep{stein2005neuronal}. 
Similarly, the human visual system can function over many orders of magnitude of objective brightness which dramatically alters the variance of the visual input, while in auditory perception the variance of specific sound inputs is crucially dependent on ambient audio conditions. In all cases, being able to represent the variance of the incoming sensory data is likely crucial to being able to successfully model and perform accurate inference on such sensory streams. 

Precision also has deep relevance to machine learning. As noted in the backpropagation section later in the paper, predictive coding with fixed predictions and identity precisions forms a scheme which can converge to the exact gradients computed by the backpropagation of error algorithm. Importantly, however, when precisions are included in the scheme, predictive coding forms a superset of backpropagation which allows it to weight gradients by their intrinsic variance. This more subtle and nuanced approach may prove more adaptable and robust than standard backpropagation, which implicitly assumes that all data-points in a dataset are of equal value - an assumption which is likely not met for many datasets with heteroscedastic noise. Exploring the use of precision to explicitly model the intrinsic variance of data is an exciting area for future work, especially applied to large-scale modern machine learning systems. Indeed, it can be shown (see Appendix B) that in the linear case using learnt precisions is equivalent to a natural gradients algorithm \cite{amari1995information}. Natural gradients modulate the gradient vector with the Hessian or curvature of the loss function (which is also the Fisher information) and therefore effectively derive an optimal adaptive learning rate for the descent, which has been found to improve optimization performance, albeit at a sometimes substantial computational cost of explicitly computing and materializing the Fisher information matrix.

There remains an intrinsic tension, however, between these two perspectives on precision in the literature. The first interprets precision as a bottom-up `objective' measure of the intrinsic variance in the sensory data and then, deeper in the hierarchy, the intrinsic variance of activities at later processing stages. This contrasts strongly with views of precision as serving a general purpose \emph{adaptive} modulatory function as in attention. While attention is indeed deeply affected by bottom up factors, which are generally termed attentional salience \citep{parkhurst2002modeling}, these factors are typically modelled as Bayesian surprise \citep{itti2009bayesian}.     `Bayesian surprise' is often modelled mathematically as the information gain upon observing a stimulus which is not necessarily the same as high or low variance. For instance, both high variance (such as strobe-lights or white noise) and low variance (such as constant bright incongruous blocks of colour) may both be extremely attentionally salient in visual input while having opposite effects on precision. Precisions, if updated using Equation \ref{Precision_update} explicitly represent the objective variance of the stimulus or data, and therefore cannot easily account for the well-documented top-down or contextually guided attention \citep{torralba2003statistics,kanan2009sun,henderson2017gaze}. This means that it seems likely that theories of top-down modulatory precision cannot simply rely on a direct derivation of precision updating in terms of a gradient descent on the free energy, but must instead postulate additional mechanisms which implement this top down precision modulation explicitly. One possible way to do this is to assume a system of direct inference over precisions with modulatory `precision expectations' which form hyperpriors over the precisions, which can then be updated in a Bayesian fashion by using the objective variance of the data as the likelihood. However, much remains to be worked out as to the precise mathematics of this scheme.

Finally, there remains an issue of timescale. Precisions are often conceptualised as being optimized over a slow timescale, comparable with the optimisation of synaptic weights -- i.e., in the M-step of the EM algorithm -- which fits their mathematical expression as a parameter of the generative model. However, attentional modulation can be very rapid, and likely cannot be encoded through synaptic weights. These concerns make any direct identification of precision with attention difficult, while the idea of precision as instead encoding some base level of variance-normalization or variance weighting finds more support from the mathematics. However even here problems remain due to timescales. The objective variance of different regions of the sensory stream can also vary rapidly, and it is not clear that this variation can be encoded into synaptic weights either, although it is definitely possible to maintain a moving \emph{average} of the variance through the lateral synaptic weights. Overall, the precise neurophysiological and mathematical meaning and function of precision remains quite uncertain, and is thus an exciting area of future development for the predictive coding framework. Finally, there has been relatively little empirical work on studying the effects of learnable precision in large-scale predictive coding networks.

\subsection{Predictive Coding in the Brain?}

While technically predictive coding is simply a variational inference and filtering algorithm under Gaussian assumptions, from the beginning it has been claimed to be a biologically plausible theory of cortical computation, and the literature has consistently drawn close connections between the theory and potential computations that may be performed in brains. For instance, the Rao and Ballard model explicitly claims to model the early visual cortex, while \citet{friston2005theory} explicitly proposed predictive coding as a general theory of cortical computation. In this section, we review work which has began translating the mathematical formalism into neurophysiological detail, and focus especially on the seminal cortical microcircuit model by \citet{bastos2012canonical}. We also briefly review empirical work that has attempted to verify or falsify key tenets of predictive coding in the cortex, and discuss the methodological or algorithmic difficulties with this approach. 

The hierarchical generative models generally treated in predictive coding are composed of multiple layers in a stacked structure. Each layer consists of a single vector of value, or activity, neurons and a single vector of error neurons. However, the cortex is not organised into such a simple structure. Instead each cortical `area' such as V1, V2, or V4 is comprised of 6 internal layers: L1-L6. These layers are reciprocally connected with each other in a complex way which has not yet been fully elucidated, and may subtly vary between cortical regions and across species \citep{felleman1991distributed}. Nevertheless, there is convergence around a relatively simple scheme where the six cortical layers can be decomposed into an `input layer' L4, which primarily receives driving excitatory inputs from the area below as well as from the thalamus, and then two relatively distinct processing streams -- a feedforward superficial, or supragranular stream consisting of layers L1/2/3 and a feedback deep, or infragranular stream consisting of layers 5 and 6 (layer 4 is typically called the `granular' layer). These streams have been shown to have different preferred oscillatory frequencies, with the superficial layers possessing the strongest theta and gamma power (\citep{bastos2015visual}, and the deep layers possessing strongest alpha and beta power which are negatively correlated across layers. The superficial layers then send excitatory connectivity forward to L4 of the area above, while the deep layers possess feedback connectivity, which can be both inhibitory or excitatory, back to both deep and superficial layers of the areas below. Within each cortical area, there is a well-established 3 step feedback relay, from the input L4, to the superficial layers L2/3, which then project their input forwards to the next area in the hierarchy. From L2/L3, the superficial layers then project to the deep Layer 5, which could then project to L6, or else provide feedback to regions lower in the hierarchy \citep{rockland2019we}. Interestingly, deep L5 and L6 are  the only cortical layers which contains neurons which project to subcortical regions or the brainstem, and L6 especially appears to maintain precise reciprocal connectivity with the thalamus \citep{thomson2010neocortical}. While this feedforward input, superficial, deep `relay' is well studied, there are also other pathways, including from deep to L4 \citep{amorim2010whose}, and superficial feedback connections which are not well explored. Moreover, alongside the cortico-cortico connectivity studied here, there are also many cortico-subcortico, and especially cortico-thalamic connections which are less well-understood or integrated into specific process theories of predictive coding \citep{markov2014anatomy}. 

\begin{figure}
    \centering
    \includegraphics[scale=0.2]{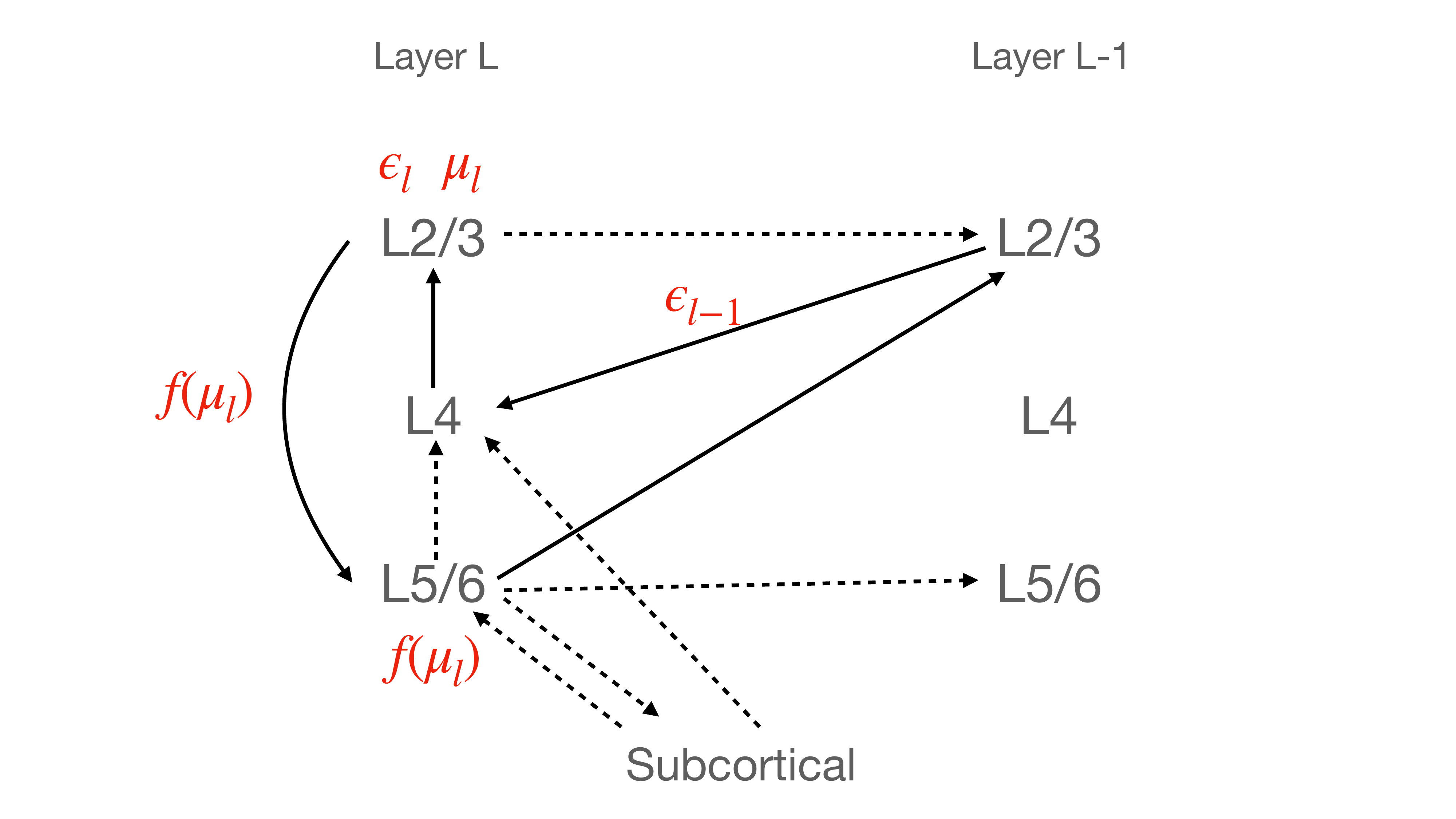}
    \caption{The canonical microcircuit proposed by Bastos et al mapped onto the laminar connectivity of a cortical region (which comprises 6 layers). Here, for simplicity, we group layers L2 and L3 together into a broad `superficial' layer and L5 and L6 together into a `deep' layer. We ignore L1 entirely since there are few neurons there and they are not involved in the Bastos microcircuit. Bold lines are included in the canonoical microcircuit of Bastos et al. Dashed lines are connections which are known to exist in the cortex which are not explained by the model. Red text denotes the values which are computed in each part of the canonical microcircuit}
    \label{bastos_circuit_figure}
\end{figure}

While this intrinsic connectivity of the cortical region may seem dauntingly complex, much progress has been made within the last decade of fitting predictive coding models to this neurophysiology. Of special importance is the work of \citet{bastos2012canonical} who provided the central microcircuit model of predictive coding \footnote{Perhaps the first worked out canonical microcircuit for predictive coding, although not using that name, was in the early work of \citet{mumford1992computational}. He argued that the descending deep pathway transmits `templates' backwards which are then fitted to the data present in the layers below before computing `residuals' which are transmitted upwards on the superficial to L4 ascending pathway.}. The fundamental operations of predictive coding require predictions to be sent down the hierarchy, while prediction errors are sent upwards. The dynamics of the value neurons $\mu$s require both the prediction errors at the current layer (from the top down predictions of the layer above) to be combined with the prediction errors from the layer below mapped through the backwards weights $\theta^T$ and the derivative of the activation function $\frac{\partial f}{\partial \mu}$. The Bastos microcircuit model associates a `layer' of predictive coding, with a 6-level cortical `region'. The inputs to L4 of the region are taken to be the prediction errors of the region below, which are then immediately passed upwards to the superficial levels L2/L3 where the prediction error $\epsilon_l$ and the value neurons $\epsilon_l$ are taken to be located. The predictions $f(\mu_l, \theta_l)$ are taken to reside in the deep layers L5/6. The superficial layers receive top-down prediction inputs from the deep layers of the region above it in the hierarchy $f(\mu_{l+1}, \theta_{l+1})$ which are combined to compute the prediction errors $\epsilon_l$ of the region. These are then combined with the bottom-up prediction errors $\epsilon_{l+1}$ coming from L4 to update the value neurons $\mu_l$, also located in the superficial layers. The value neurons then transmit to the deep layers L5/6 where the predictions $f(\mu_l, \theta_l)$ to be transmitted to the region below on the hierarchy are computed, while the superficial prediction error units $\epsilon_l$ transmit to the L4 input layer of the region above in the hierarchy.  The full schematic of the Bastos model is presented in Figure \ref{bastos_circuit_figure}. 

This model fits the predictive coding architecture to the laminar structure within cortical regions. It explains several core features of the cortical microcircuit -- that superficial cells (interpreted as encoding prediction errors), project forward to the L4 input layer of the region above. It also provides an interpretation of the function of the well-established `relay' from L4 to superficial (transmitting inputted prediction errors to superficial), the computation of prediction errors and value neurons in the superficial layers, and then the superficial to deep connectivity as encoding the prediction of the value neurons. It also can explain the deep to superficial backwards pass as the transmission of predictions from one region to the next. Similarly, the L5 and L6 deep cortico-subcortical transmission can be straightforwardly interpreted as the transmission of predictions to subcortical areas, or to motor and brainstem regions to perform active inference. However, there are several aspects of the model and the neurophysiology which require further elucidation. One primary issue is the well-established deep-deep feedback pathway from the deep layers of one region to the deep layers of the region below it in the hierarchy \citep{amorim2010whose,harris2018organization}. In strength, this feedback pathway is often considered more important than the deep-superficial pathway that is thought to convey the predictions to the prediction error units in the superficial layers of the region below. This pathway is entirely unexplained within the Bastos model, yet it appears to be important for cortical function. It is possible that this may be thought of as a prediction pathway, so that predictions at one level can directly modulate predictions at the layer below without having to have their influence modulated by going through the prediction error units. This would provide the brain with a powerful downward or generative path, enabling it to compute low-level predictions effectively directly from high-level abstractions in a single downward sweep. However, such a path is not used in the mathematical predictive coding model, and it is unclear what probabilistic interpretation it could have. A secondary concern is the fact that there also exists a superficial-superficial feedback connection with unclear function within this model \citep{markov2014anatomy}. It has been suggested that this feedback connection may carry precision information \citep{kanai2015cerebral}, although it is unclear why this is necessary since the actual dynamics of the precision only require access to the precision-weighted prediction errors of the current layer (Equation \ref{Precision_update}).

A more general concern with this model is that the deep layers are in general relatively poorly utilised by the model. All the `action' so to speak occurs in the superficial layers -- which is where both the prediction errors and the value neurons are located, and where the top-down predictions interface with the prediction errors. The only task the deep layers provide in this model is to compute the predictions and then relay them to the layers below. It is possible, perhaps likely, that the actual function $f(\theta, ;\mu)$ could be considerably more complex than the ANN approach of synaptic weights passed through an elementwise nonlinear activation function. However, extensions of predictive coding to more complex backwards functions have not yet been substantially explored in the literature, and would require more complex update rules for the parameters which may lead to less biologically plausible learning rules. It is also known that the deep layers contain the connections to the thalamus, striatum, and other subcortical regions which are likely important in action selection as well as large-scale coordination across the brain. However, such effects and connections are not included in standard predictive coding models which are primarily concerned with only cortico-cortical processing.

One interesting potential issue is that there are several null operations in the model. For instance, the prediction errors $\epsilon_{l-1}$ are computed in the superficial layers at the level below, and then transmitted first to the input layer L4 before being transmitted again to the superficial layers. This provides two sets of operations on the prediction errors while only one is necessary, thus necessitating that one of these steps is effectively a pure relay step without modification of the prediction errors -- an interesting and testable neurophysiological prediction. A similar situation arises with the predictions, where although the predictions are considered to only be computed as a nonlinear function of the value units, mapped through synaptic connections, the predictions actually undergo two steps of computation. Firstly in the superficial-deep transmission within a region, and secondly the deep-superficial feedback transmission to the region below. According to the standard model, one of these steps must be a null step and not change the predictions, which could in theory be tested by current methods. Importantly, it is possible that the actual function being computed by the predictions is more complex (although not the prediction errors), and thus takes multiple steps. However, to achieve learning in such a system would require more complicated update rules, which would likely exacerbate the issues of weight transport and derivative computation already inherent in the algorithm. An additional interesting consideration is the extension of predictive coding from a simple linear stack of hierarchical regions, to a heterarchical scheme, where multiple regions may all project to the same region, and similarly one region may send prediction errors to many others. Predictive coding makes a very strong hypothesis in this situation, which is that heterarchical connectivity must be symmetrical. If a region sends feedforward prediction errors to another region, it must receive feedback predictions from it and vice versa. This feature of connectivity in the brain has long been confirmed through neuroantomical studies \cite{mumford1992computational,felleman1991distributed}.

While the prediction errors must be transmitted upwards by being modulated through the transpose of the forward weights $\theta^T$, which would be implemented as the synaptic strengths in either the deep-superficial backwards paths, or the superficial-deep forward relay step, both of which are far from the superficial-input pathway, thus raising a considerable issue of weight transport.  Additionally, these prediction errors must transmit with them the derivative of the activation function $\frac{\partial f}{\partial \mu}$, which is theoretically available at the superficial layers of the level below where the prediction errors are transmitted, but would then need to be computed separately and transmitted back up with the prediction errors. The weight transport problem poses a greater difficulty, however, as we discuss below it may be solvable with learnable or random backwards weights.

A further potential issue arises from the interplay of excitation and inhibition within the microcircuit \citep{kogo2015predictive,heilbron2018great}. Specifically, predictive coding requires the feedback connectivity containing predictions to be inhibitory, or else it requires the connection between the error and value neurons to be inhibitory, depending on the direction of the update. However, both cortico-cortico feedback projections as well as pyramidal-pyramidal interactions within laminar layers both tend to be excitatory. To address this, additional inhibitory interneuron circuitry may be required to negate one of these terms which remains to be explored. For instance, \citet{shipp2016neural} suggests that the Martinotti cells could perform this function. An additional consideration is the fact that while the mathematical form of predictive coding allows for negative prediction errors, these cannot be implemented directly in the brain -- a neuron cannot have a negative firing rate. Negative firing rates could be mimicked by using a high baseline firing rate and then interpreting firing below the baseline to be negative, although to have a baseline high enough to enable a roughly equal amount of positive and negative `space' for encoding would be extremely energetically inefficient \citep{keller2018predictive}. Another potential solution, as suggested in \citet{keller2018predictive} would be to use separate populations of positive and negative prediction errors, although then precise circuitry would be needed to ensure that information is routed to the correct positive or negative neuron, or additionally that each value neuron would have to stand in a one-to-one relationship with both a positive and negative error neuron, with one connection being inhibitory and the other being excitatory, and it is unclear whether such precise connectivity can exist in the brain. Finally, prediction error neurons could encode their errors in an antagonistic fashion, as do color-sensitive opponent cells in the retina -- for instance a red-green opponent cell could simply encode negative green prediction error as positive red prediction error. However, it is unclear to what extent opponent cells exist in deeper, more abstract regions of the cortex, nor what the opponency would signify \citep{huang2011predictive}.

The implementation of precision in such a predictive coding microcircuit is another interesting question which has yet to be fully fleshed out. \citet{kanai2015cerebral} suggests that precision may be encoded in subcortical regions such as the pulvinar, which is known to be engaged in attentional regulation. They suggest that such precision modulation  could be implemented either as direct modulation of the superficial error units by neurons projecting from the pulvinar, or else alternatively via indirect effects of the pulvinar being instrumental in establishing or dissolving synchrony within or between regions, which is known to affect the amount of information transmitted between layers \citep{kanai2015cerebral,buzsaki2006rhythms,uhlhaas2006neural}. However, mathematically, precision is a matrix which quantifies the covariance between each error unit and each other error unit within the layer. It is unlikely that the pulvinar could usefully process or precisely modulate this $N^2$ amount of information. The pulvinar could, however, be instrumental in computing a diagonal approximation to the full precision matrix, by essentially modulating each superficial error neuron independently while lateral connectivity within the layer, perhaps mediated by SST interneurons, which are known to project relatively uniformly to a local region, could be involved in the implementation of the full precision matrix. The pulvinar could potentially focus on global exogenous precision modulation, such as due to attention, while the lateral inhibition would focus primarily on modelling the bottom-up aspects of precision such as the prediction error variance which arises naturally from Equation \ref{Precision_update}. 
If such a scheme were implemented in the brain, with diagonal global precision modulation, and full-matrix lateral precision computation, then this immediately suggests the intriguing hypothesis that top-down attention can only modulate independent variations in stimulus aspects, while bottom-up attention or salience can and does explicitly model their covariances.

One final interesting consideration arises from the consistent and well-established finding that the superficial and deep layers operate at different principal frequencies, with the superficial layers operating at the fast gamma frequencies, while the deep layers primarily utilize the slower alpha and beta frequencies \citep{bastos2015visual}. This finding does not necessarily follow from the cortical microcircuit model above, which if anything suggests that predictions and prediction errors, and thus superficial and deep layers should operate at roughly the same frequency. It has been argued \citep{bastos2012canonical}, that the predictions could operate at a slower frequency, since they integrate information from the prediction errors over time, however this would be an additional assumption, not a direct consequence of the standard mathematical model, in which the predictions are an instantaneous function of the value neurons. Now, the value neurons themselves, since they are updated using iterative dynamics, \emph{do} integrate information from the instantaneous prediction errors, and thus would potentially have a slower frequency, however the value neurons are intermingled with the prediction error neurons in the superficial layers, and would thus also be expected to affect the dominant frequency of the superficial layers. It is possible, however, that the higher frequency prediction error neurons operating at the gamma frequency disguise the lower frequency value neurons in their midst, while the alpha/beta signal of the deep layers, which only contain prediction neurons, is not so disguised, giving rise to the observed frequency dynamics.

Overall, while much progress has been made in translating the abstract mathematical specification of predictive coding into neurophysiologically realistic neural circuitry, there are still many open questions and important problems. The fit to the cortical microcircuitry is not perfect, and there are several known cortical pathways which are hard to explain under current models. Nevertheless, predictive coding process theories provide perhaps some of the clearest and most general neuronal process theories relating cortical microcircuitry to an abstract computational model which is known to be able to solve challenging cognitive tasks. Finally, all the process models considered have assumed that neurons primarily communicate through real-valued rate-codes instead of spiking codes. If the brain does use spiking-codes heavily, then the algorithmic theory and process theories of predictive coding would need to be reformulated, if possible, to be able to natively handle spiking neural networks. In general, learning and inference in spiking neural network models remains relatively poorly understood, although there has been much recent progress in this area \citep{zenke2018superspike,bellec2020solution,kaiser2020synaptic,neftci2019surrogate,zenke2021remarkable}. The extension of predictive coding-like algorithms to the spiking paradigm is also an exciting open area of research \citep{ororbia2019biologically,boerlin2013predictive,brendel2020learning}. An additional complication, which may lead to novel algorithms and implementations is the fact that neurons have several distinct sites of dendritic integration \citep{sacramento2018dendritic, takahashi2020active}, as well as complicated internal, synaptic, and axonal physiology which may substantially affect processing or offer considerably more expressive power than the current understanding of `biological plausibility' admits. 

There has also been a substantial amount of research empirically investigating many of the predictions made by the process theories of predictive coding \citep{bastos2012canonical,keller2018predictive,kanai2015cerebral}. A recent thorough review of this work is \citet{walsh2020evaluating}. A large amount of research has focused on the crucial prediction that expected, or predicted, stimuli should elicit less error response than unexpected ones. While the neural phenomena of repetition-suppression and expectation-suppression are well-established at the individual unit level, these phenomena are also well explained by other competing theories such as neural adaptation \citep{desimone1995neural}. Evidence from large-scale fMRI studies of the activity of whole brain regions are mixed, with some studies finding increases and others decreases in activity. Additionally, predictive coding does not actually make clear predictions of the level of activity in whole brain regions. While predictive coding predicts that the activity of error neurons should drop, the error neurons are generally thought of as being situated in the superficial layers alongside the value neurons, whose activity may rise. Additionally, well-predicted stimuli might be expected to have high precision, which would then boost the error unit activity, thus counteracting to some extent the drop due to better prediction. Precisely how these effects would interact in aggregate measures like the fMRI BOLD signal is unclear.

Another approach to empirically testing the theory is to look for specific value and error units in the brain. However, this task is complicated by the fact that often the predictions against which the errors are computed are unknown. For instance, it is well known that there are neurons in V1 which are sensitive to the illusory contours of stimuli such as the Kanizsa triangle \citep{kanizsa1955margini,kok2015predictive,kok2015role}, however, there remains a problem of interpretation. It is not clear whether such neurons are prediction errors, since there was a prediction of the contour which was not in fact there, or whether they represent value neurons faithfully representing the prediction itself. Nevertheless, there has been some evidence of functionally distinct neuronal subpopulations potentially corresponding to prediction and prediction error neurons \citep{bell2016encoding,fiser2016experience}.


It is also important to note that some implementations of predictive coding do not necessarily require separate populations of error units, but instead assume multicompartmental neurons with a distinct apical dendrite compartment which can store prediction errors separately from the value encoded by the main neuron body \citep{sacramento2018dendritic,takahashi2020active} \footnote{While the original architecture of \citet{sacramento2018dendritic} was not explicitly derived from predictive coding, it has later been shown that the two are equivalent \citep{whittington2019theories}.}. If predictive coding were to be implemented in the brain in such a fashion, then not finding explicit error units would not conclusively refute predictive coding. Due to this flexibility in both the theory, and the process theory translating it to neural circuitry, as well as the difficulty in extracting predictions of aggregate measures (such as fMRI BOLD signal) from the mathematical model, it has been challenging to either experimentally confirm or refute predictive coding as a theory up to now. However, with emerging advances in experimental techniques and methodologies, as well as theoretical progress in exploring the landscape and computational efficacy of different predictive coding variants, as well as making more precise process theories, predictive coding, or at least specific process-theories, may well be amenable to a definitive experimental verification or falsification in the future.

\section{Paradigms of Predictive Coding}

While predictive coding has quite a straightforward mathematical form, there are numerous ways to set up predictive coding networks to achieve particular tasks, and numerous subtleties which can hinder a naive implementation. In this section, we survey the different paradigms of training predictive coding networks and review the empirical studies which have been carried out using these types of network. In brief, we argue that predictive coding can be instantiated in a supervised or unsupervised fashion. In the unsupervised case, there is a hierarchy of nodes, but the top-level of the hierarchy is allowed to vary freely. New data enters only at the bottom level of the hierarchy. In this case, the predictive coding network functions much like an autoencoder network \citep{hinton1994autoencoders,kingma2013auto}, in which its prediction is ultimately of the same type of its input. In the supervised case, the activity nodes at the highest latent level of the network are fixed to the values given by the supervisory signal -- i.e. the labels. The supervised network can then be run in two directions, depending on whether the `data' or the `label' is provided to the top or the bottom of the network respectively. 

In the following, we review empirical work demonstrating the performance characteristics of predictive coding networks in both unsupervised and supervised scenarios. We then summarize work in making the standard predictive coding architecture more biologically plausible by relaxing certain assumptions implicit in the canonical model of predictive coding as described so far, as well as how predictive coding networks can be extended to perform \emph{action}, through active inference.

\begin{figure}
    \centering
    \includegraphics[scale=0.2]{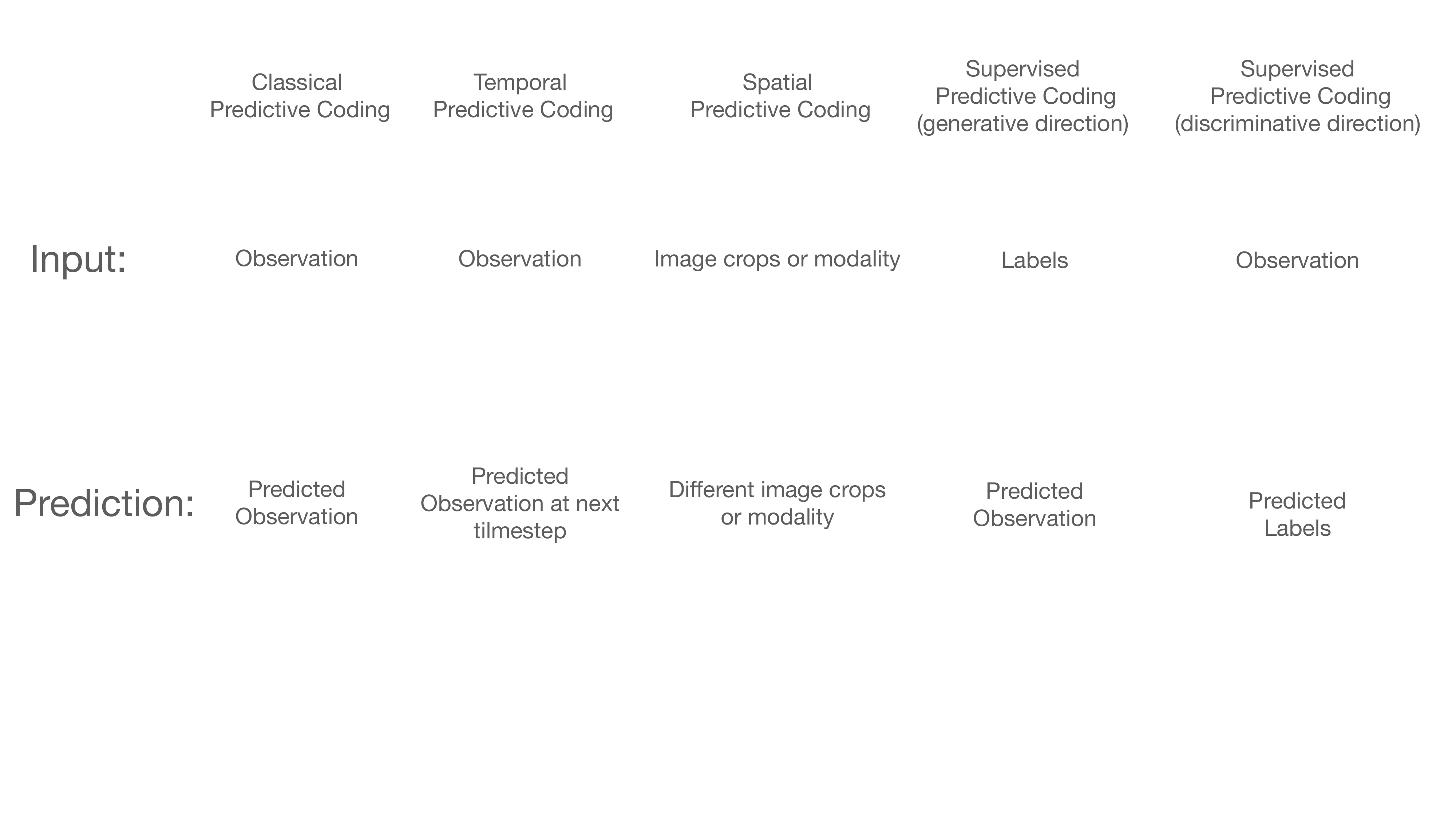}
    \caption{Summary of the input output relationships for each paradigm of predictive coding. Specifically a.) What the input to the network is and b.) what the network is trained to predict.}
    \label{fig:my_label}
\end{figure}

\subsection{Unsupervised predictive coding}

Unsupervised training is perhaps the most intuitive way to think about predictive coding, and is the most obvious candidate for how predictive coding may be implemented in neural circuitry. On this view, the predictive coding networks functions essentially as an autoencoder \citep{hinton1994autoencoders,kingma2013auto,hinton2006reducing}, attempting to predict either the current sensory input, or the next 'frame' of sensory inputs (temporal predictive coding).  Under this model the latent activations of the highest level are not fixed, but can vary freely to best model the data. In this unsupervised case, the question becomes what to predict, to which there are many potential answers. We review some possibilities here, which have been investigated in the literature. 

To train an unsupervised predictive coding network, the activities of the lowest layer are fixed to those of the input data. The activities of the latent variables at all other levels of the hierarchy are initialized randomly. Then, an iterative process begins in which each layer makes predictions at the layer below, which generates prediction errors. The latent variable $\mu$s then follow the dynamics of equation (PC) to minimize prediction errors. Once the network has settled into an equilibrium, or else the dynamics have been run for some prespecified number of steps, then the parameters $\theta$ are updated for one step using the dynamics of Equation \ref{PC_update_rules}. If the precisions are learnt as well, then the precision dynamics (Equation \ref{Precision_update} will be ran for one step here as well. To help reduce the variance of the update, predictive coding networks are often trained with a minibatch of data for each update, as in machine learning \citep{friston2005theory,whittington2017approximation,millidge2019implementing,millidge2019combining,orchard2019making,millidge2018predictive}. In general, predictive coding networks possess many of the same hyperparameters such as the batch size, the learning rate, and layer width as artificial neural network models from machine learning. Predictive coding networks can even be adapted to use convolutional or recurrent architectures \citep{millidge2020predictive,salvatori2021predictive}. The main difference is the training algorithm of predictive coding (Equations \ref{PC_update_rules}) rather than stochastic gradient descent with backpropagated gradients, although under certain conditions it has been shown that the predictive coding algorithm approximates the backpropagation of error algorithm \citep{millidge2020predictive,whittington2017approximation,song2020can}.

\subsubsection{Autoencoding and predictive coding}

A simple implementation of predictive coding is as an autoencoder \citep{hinton2006reducing}. A data item $o$ is presented to the network, and the network's goal is to predict that same data item $\hat{o}$ back from the network. The goal of this kind of unsupervised autoencoding objective is typically to learn powerful, structured, compressed, or disentangled representations of the input.
In a completely unconstrained network, the solution is trivial since the predictive coding network can just fashion itself into the identity function. However, by adding constraints into the network this solution can be ruled out, ideally requiring the network to learn some other, more useful representation of the input to allow for correct reconstruction. Typically, these constraints arise from a physical or informational bottleneck in the network \citep{tishby2000information} -- typically by constricting the dimensionality of one or multiple layers to be smaller (often significantly smaller) than the input dimensionality, thus effectively forcing the input to be compressed in a way amenable to later decompression and reconstruction. The activations in the smallest bottleneck layer are then taken to reflect important representations learned by the network. Autoencoders of this kind of have been widely used in machine learning, and variational autoencoders \citep{kingma2013auto}, a probabilistic variant which learns a Gaussian latent state, are often state of the art for various image generation tasks \citep{child2020very}. Such networks can be thought of as having an `hourglass shape', with wide encoders and decoders at each end and a bottleneck in the middle which encodes the latent compressed code. Predictive coding networks, by contrast, only have the bottom half of the hourglass (the decoder), since the latent bottleneck states are optimized by an iterative inference procedure (the E step of the E-M algorithm).

An early example of an autoencoding predictive coding network was provided by \cite{rao1999predictive}, who showed that such a network could learn interpretable representations in its intermediate layers. The representation learning capabilities of predictive coding autoencoder networks has also been tested on the standard machine learning dataset of MNIST digits \citep{millidge2019implementing}, where it was shown that the latent code can be linearly separated by PCA into distinct groups for each of the 10 digits, even though the digits are learned in an entirely unsupervised way. 

\subsubsection{Temporal Predictive Coding}

Another way to implement an unsupervised predictive coding network is in a temporally autoregressive paradigm \citep{spratling2010predictive,friston2008hierarchical,millidge2019fixational}. Here, the network is given a time-series of data-points to learn, and it must predict the next input given the current input and potentially some window of past inputs. This learning objective has additional neurobiological relevance (as compared to simple autoencoding), given that the brain is in-practice exposed to continuous-time sensory streams, in which predictions must necessarily be made in this temporally structured way. 

Furthermore, predicting temporal sequences is a fundamentally more challenging task than simply reconstructing the input, since the future is not necessarily known or reconstructable given the past, and a simple identity mapping will not suffice except for the most trivial sequences. Indeed, autoregressive objectives like this have been used in machine learning to successfully train extremely impressive and large transformer models to predict natural language text to an incredibly high degree of accuracy \citep{vaswani2017attention,brown2020language}. Autoregressive predictive coding networks have primarily been explored in the context of signal deconvolution in neuroimaging \citep{friston2008DEM,friston2008hierarchical}, as well as in predictive coding inspired machine learning architectures such as PredNet \citep{lotter2016deep}. 

Such networks have also been used in the context of reinforcement learning for 1-step environment prediction to enable simple planning and action selection \citep{millidge2019combining}. Moreover, as shown previously, 1-step autoregressive linear predictive coding is mathematically similar to Kalman Filtering \citep{millidge2021neural, friston2008hierarchical}, thus demonstrating the close connection between predictive coding and standard filtering algorithms in engineering. In general, however, despite the empirical successes and importance of unsupervised autoregressive modelling in machine learning, and its neurobiological relevance, surprisingly little work has been done on empirically testing the abilities of large-scale predictive coding networks on autoregressive tasks. 

\subsubsection{Spatial Predictive Coding}

Another objective which could be used in predictive coding is to predict parts of the input from other parts. Specifically, it is possible to get a predictive coding network to learn to predict a pixel or spatial element from its surroundings. This objective has been used in early work on predictive coding in the retina \citep{srinivasan1982predictive} which models the receptive field properties of retinal ganglion cells. This paradigm of predictive coding has close relationships to normalization or whitening transforms. A closely related paradigm is cross-modal predictive coding, which uses information from one sensory modality to predict another. This has been explored in \citep{millidge2018predictive} and has been shown to lead to good representation learning performance with cross-predicting autoencoders which cross-predict using the three colour channels (red,green,blue) of natural images. The brain may use a cross-modal predictive objective more widely, as suggested by the close integration of multimodal inputs and the ability to effortlessly make cross-modal predictions.


A similar approach is taken in machine learning where it is called contrastive predictive coding \citep{oord2018representation}, which aims to learn latent representations by forcing the network to be able to associate two different crops of the same image, while dissociating crops of different images. This contrastive approach has demonstrated strong results on unsupservised machine learning benchmarks

\subsection{Supervised predictive coding: Forwards and Backwards}

Supervised predictive coding is the second major paradigm of predictive coding. In the supervised mode, both data and supervisory labels are provided to the network. The training objective is to learn some function of the data that will allow the network to successfully predict the correct labels. In supervised predictive coding, the activities of the units at one end of the network is fixed to the data, and the other end is fixed to the values of the labels. Then, predictions are computed from the latent variables at the top of the network and fed down to the bottom, generating prediction errors at each step. The activations of all intermediate $\mu$s are then allowed to evolve according to the dynamics of Equation \ref{PC_update_rules} while the activations at the top and bottom layers of the network are fixed to the label or the data values. Once the network has settled into an equilibrium, then the parameters of the network are updated. This is repeated for each minibatch of data points and labels. 

There are two separate modes for running the predictive coding network in the supervised case -- the forward mode and the backwards mode. The forward mode is more intuitive, in which $\mu$s at the top of the hierarchy are fixed to the labels while the bottom of the network is fixed to the data values. Predictions thus flow down the hierarchy from the labels to the data. To test the network, the bottom nodes of the network are fixed to the test data item, and the label latent variables are allowed to freely evolve according to Equation \ref{PC_update_rules}. The label prediction the network makes is determined by the values the top layer of $\mu$s have taken on after convergence. Thus, in the forward mode convergence is an iterative process requiring multiple dynamical iterations. In the backwards mode, the bottom of the network is fixed to the labels, and the top of the network is fixed to the datapoint. Predictions thus flow directly from data to labels in a manner reminiscent of the feedforward pass of an artificial neural network. In this case, at test time, all the network needs to do is to make a single forward (really downward) pass from data to labels without requiring multiple dynamical iterations. We will show later that this backwards predictive coding network can become equivalent to backpropagation of error in artificial neural networks under certain conditions, which provides a revealing link between predictive coding and contemporary machine learning.

\begin{figure}%
    \centering
    \subfloat[\centering `Standard' Generative PC]{{\includegraphics[width=0.47\textwidth]{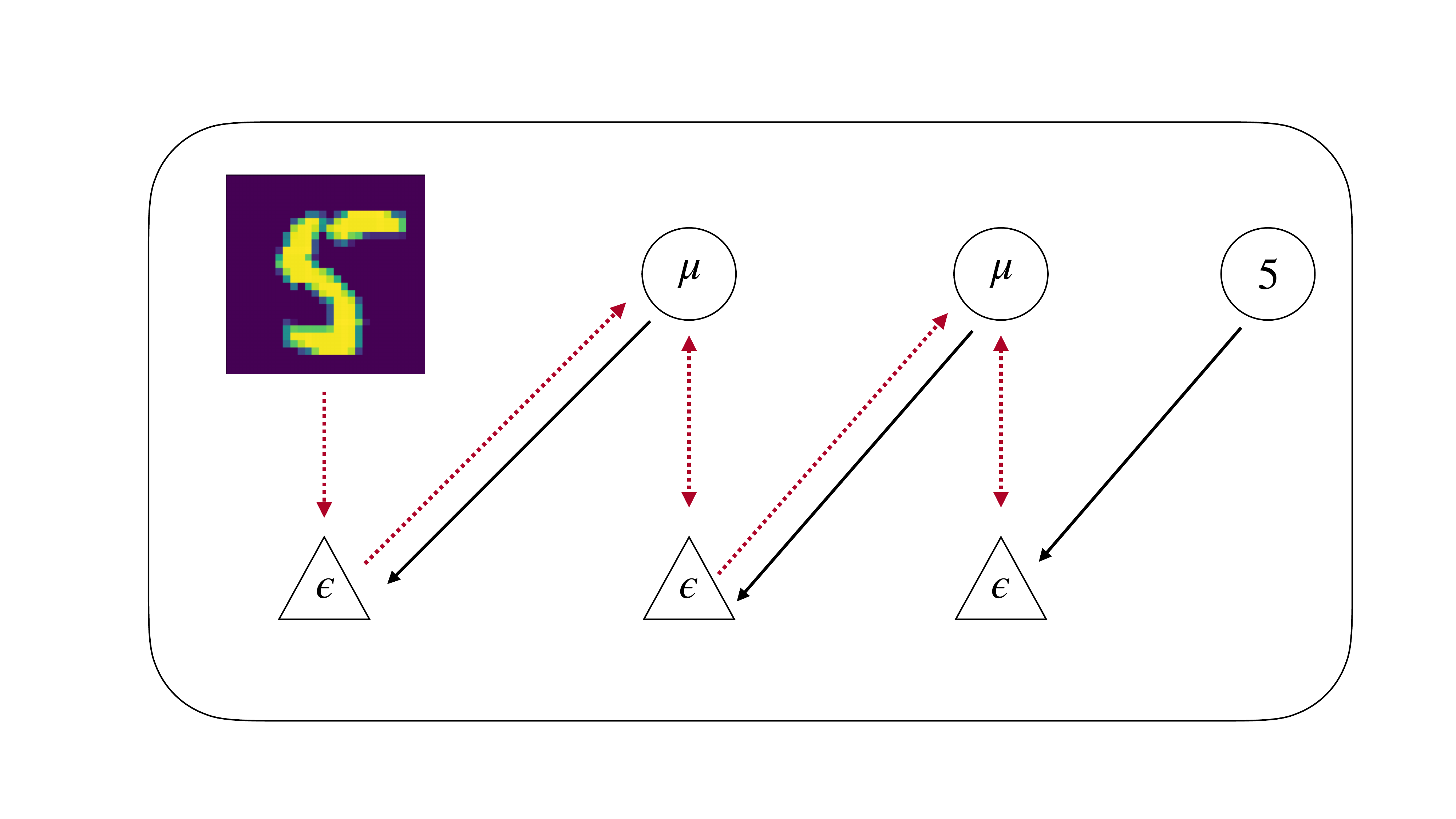} }}%
    \qquad
    \subfloat[\centering `Reverse' Discriminative PC ]{{\includegraphics[width=0.47\textwidth]{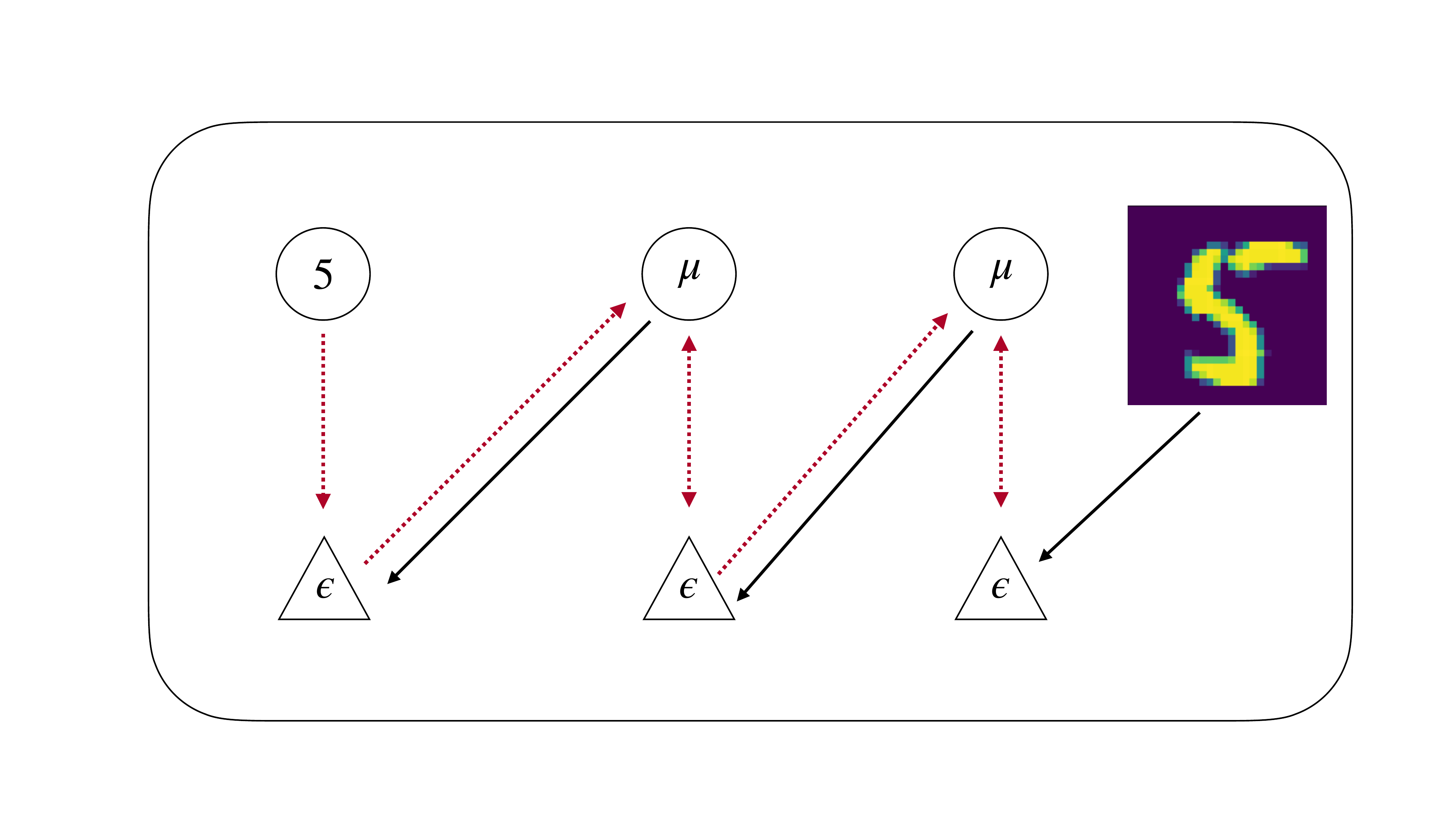} }}%
    \caption{Schematic architectures for the a.) Standard, or generative predictive coding setup, or b.) Reverse, or discriminative architecture trained for supervised classification on MNIST digits. In the generative model, the image input (in this case an MNIST digit) is presented to the bottom layer of the network, and the top layer is fixed to the label value (5). Predictions (in black) are passed down and prediction errors (in red) are passed upwards until the network equilibrates. In the discriminative mode, the input image is presented to the top of the network and the label is presented at the bottom. Thus the network aims to `generate' the label from the image. The top-down flow of predictions becomes analogous to the forward pass in an artificial neural networks, and the bottom-up prediction errors become equivalent to the backpropagated gradients.}%
\end{figure}

The forwards/backwards distinction in supervised predictive coding  also maps closely to a distinction between generative and discriminative classifiers in machine learning \citep{bouchard2004tradeoff}. While both forward and backwards predictive coding networks can perform both generation of `imagined' input data given labels, as well as classification (prediction of the labels given data), in each mode one direction is `easy' and the other is `hard'. The easy direction requires only a single  sweep of predictions to generate the relevant quantity -- labels or data -- while the hard direction requires a full set of dynamical iterations to convergence to make a prediction of either the labels or data. In forward predictive coding networks, generation is easy, while classification is hard. Predictions flow directly from the labels to the data, whereas to generate predictions of the labels, the network must be run until it converges. Conversely, in a backwards predictive coding network, classification is easy, requiring only a downward sweep of predictions, whereas generation requires dynamical convergence. In effect, the `downwards' sweep in predictive coding networks is always the `easy' direction, so whatever quantity is represented at the lowest level of the hierarchy will be easiest to generate. Conversely, the `upwards' direction in the predictive coding network is difficult, and thus whatever quantity is represented at the top of the hierarchy will require an iterative inference procedure to infer. In a forwards `generative mode', we have the images at the bottom and the labels at the top while in the backwards `discriminative mode', we have the labels at the bottom, and the images at the top.

In general, as might be expected, performance on easy tasks in predictive coding networks is better than performance on hard tasks. In the simple task of MNIST classification, forward predictive coding networks typically manage to generate example digits given labels with high fidelity \citep{millidge2019implementing}, while their classification accuracy, while good, is  not  comparable with artificial neural networks trained with backprop. Conversely, backwards predictive coding networks are often able to reach classification accuracies comparable to backprop-trained ANNs on MNIST, while often their generated images are blurry or otherwise poor. \citet{orchard2019making} have argued that this poor generation ability in backwards predictive coding networks arises from the fact that the generative problem is under determined -- for any given label there are many possible images which could have given rise to it -- and so the network `infers' some combination of all of them, which is generally blurry and does not correspond to a precise digit. They propose to solve this problem with a weight-decay regulariser, which encourages the network to find the minimum norm solution.

\subsection{Relaxed Predictive Coding}

Although there has been much work trying to determine the kinds of neural circuitry required to implement the predictive coding algorithm, and whether that circuitry can be mapped to that known to exist in the cortex, there are also additional problems of biological plausibility of the predictive coding algorithm which must be raised. Regardless of any details of the circuitry, there are three major implausibilities inherent in the algorithm which would trouble any circuit implementation. These are the weight transport problem, the nonlinear derivatives problem, and the error connectivity problem. The weight transport problem concerns the $\theta^T$ or backwards weight issue in the equation for the dynamics of the $\mu$s Equation \ref{PC_update_rules}. Biologically, what this term says is that the prediction errors from the layer affect the dynamics of the $\mu$ at each layer by being mapped backwards through the forward weights $\theta$ used for prediction. Taken literally, this would require the prediction errors to be transmitted `backwards' through the same axons and synapses as the predictions were transmitted forwards. In the brain, axons are only unidirectional so this is prima-facie implausible. The other option is that there is maintained a set of backwards weights which is a perfect copy of the forward weights, and that the prediction errors are mapped through these backwards weights instead. The first problem with this solution is that it requires perfectly symmetrical connectivity both forwards and backwards between layers. The second, and more serious, problem is that it requires the backwards synapses to have actually the same weight values of the forward synapses, and it is not clear how this equivalence could be initialized or maintained during learning in the brain.

The second issue of biological implausibility is the nonlinear derivative problem, or the fact that the learning rules for both $\mu$ and $\sigma$ contain the derivative of the nonlinear activation function. Although in certain cases, such as rectified linear units, this derivative is trivial, in other cases it may be challenging for neurons to compute. The third biological implausibility is that predictive coding, interpreted naively, requires a precise one-to-one connectivity between each `value neuron' $\mu$ and its `error neuron' $\epsilon$. Such a precise connectivity pattern would be hard to initialize and maintain in the brain and, if it existed, would almost certainly have been detected by current neurophysiological methods. While these three problems seem daunting for any neurobiologically plausible process theory of predictive coding in the brain, recent work \citep{millidge2020relaxing} has begun to attack these problems among others and show how the predictive coding algorithm can be relaxed to resolve these problems while still maintaining high performance.

Specifically, it has been shown \citep{millidge2020relaxing} that the weight transport problem can be addressed with a learnable set of backwards weights $\psi$ which can be initialized randomly and independently from the forward weights $\theta$. Then the backwards weights can be updated with the following Hebbian learning rule,
\begin{align*}
    \frac{d \psi_l}{dt} = \mu_{l+1} \frac{\partial f(\theta, \mu_{l+1})}{\partial \psi} \epsilon_l^T \numberthis
\end{align*}

which is just a multiplication of the (postsynaptic) latents at a layer and the (presynaptic) prediction errors of the layer below, weighted by the derivative of the activation function. \citet{millidge2020relaxing} have shown that starting with randomly initialized weights $\psi$ and learning them in parallel with the forward weights $\theta$ gives identical performance to using the correct backwards weights $\theta^T$ in supervised learning tasks on MNIST.

Similarly, the nonlinear derivative problem can, surprisingly, be handled by simply dropping the nonlinear derivative terms from the parameter and latent update rules, thus rendering the `backwards pass' of the network effectively linear. The resulting rules become,
\begin{align*}
    \frac{d\mu_l}{dt} &=  \Sigma_{l-1}^{-1} \epsilon_{l-1} \theta_{l}^T - \Sigma^{-1}_l \epsilon_l \numberthis \\
    \frac{d\theta_l}{dt} &=  \Sigma_l^{-T} \epsilon_{l-1} \mu_l^T \numberthis
\end{align*}

Surprisingly, this appears to not unduly affect performance of predictive coding networks on classification tasks. Finally, the one-to-one connectivity of latent variable nodes $\mu$ and their corresponding error units $\epsilon$ can be relaxed by feeding the predictions through an additional learnable weight matrix $\zeta$, so that the computation rule for the prediction errors becomes 
\begin{align*}
    \epsilon_l = \mu_l - \zeta_l f(\theta_l \mu_{l+1}) \numberthis
\end{align*}
neurally, the $\zeta$ would implement a fully-connected connectivity scheme between each $\mu$ and each top-down prediction. While a fixed randomly initialized $\zeta$ negatively impacts performance, the $\zeta$ matrix can also be learned online throughout the supervised learning task according to the following update rule,
\begin{align*}
    \frac{d\zeta}{dt} = \mu_l \epsilon_l^T \numberthis
\end{align*}
which was found to allow the predictive coding network to maintain performance while avoiding precise one-to-one connectivity between prediction error units and latent variables. Overall, these results show that there is in fact considerable flexibility in relaxing certain assumptions in the predictive coding update rules while maintaining performance in practice, and this relaxing of constraints opens up many more possibilities for neurobiologically accurate process theories of predictive coding to be developed and matched to neural circuitry. Moreover, it also suggests that experimental work that tries to prove or disprove predictive coding by looking for naive implementational details such as distinct prediction error neurons in one-to-one correspondence with value neurons may not prove conclusive, since there is considerably implementational flexibility of the predictive coding model into actual cortical microcircuitry.

\subsection{Deep Predictive Coding}

While so far in this review, we have considered only direct variations on the Friston, and Rao and Ballard, models of predictive coding, which are relatively pure and only use local, biologically plausible learning rules, there also exists a small literature experimenting with predictive-coding inspired deep neural networks. These are typically trained with backprop and, while they are not `pure' in that they do not faithfully implement a Rao and Ballard-esque scheme, and are not biologically plausible, by utilizing recent advances in machine learning, they often achieve substantially better performance on more challenging tasks than the purer models can achieve. As such, they provide a vital thread of evidence about the scaling properties and performances of deep and complex predictive coding networks, as might be implemented in the brain. Intuitively, deep predictive coding networks can improve upon the standard feedforward artificial neural networks which are ubiquitous in machine learning through the use of feedback connections and recurrence, which in theory allow the network to handle temporally extended tasks more naturally, as well as the feedback connections allow for an unbounded amount of computation and progressive updating or sharpening of representations and predictions over multiple recurrent iterations \citep{kietzmann2019recurrence,van2020going}

The first major work in this area is PredNet \citep{lotter2016deep}, which uses multiple layers of recurrent convolutional LSTMs to implement a deep predictive coding network. In accordance with the predictive coding framework, each convolutional LSTM cell received as input the error from the layer below. The error itself was calculated as the difference between the recurrent prediction of the error from the last timestep, as well as the top-down prediction from the layer above. An interesting quirk of this architecture was that instead of the network's goal being to predict the activation at the layer below, it was instead to \emph{predict the prediction error} at that layer. It is unclear to what extent this difference affects the behaviour or performance of PredNet against a more standard deep predictive coding implementation.  The network was trained on a number of video object recognition tasks such as a face-pose dataset and the KITTI dataset comprised of images taken from a car-mounted camera. They showed that their network outperformed baseline feedforward-only convolutional neural networks. The loss function optimized was the sum of errors at all layers, although interestingly they found that optimizing only the error at the bottom layer performed the better. The parameters of the network were optimized using backpropagation through time on the loss function. 

Although PredNet has become the seminal work in this field, it has been recently criticised for its lack of adherence to the pure predictive coding model, and related lack of biological plausibility, as well as questions over the degree to which it is learning useful representations rather than simply predicting the low-level optical flow in the images \citep{rane2020prednet}, which is always a danger in video prediction where straightforward extrapolations of optical flow can perform surprisingly well. This work has been further developed in \citet{wen2018deep} who develop a recurrent-convolutional scheme with top-down deconvolutional predictions, and demonstrate that the recurrence the network enables allows for greater performance than strictly feedforward networks, and additionally that the degree of performance increase scales with the number of recurrent iterations allowed.  For a further review of predictive-coding inspired deep learning architectures, see \citet{hosseini2020hierarchical}

Moving further into the field of machine learning, there is also a number of papers which utilize a predictive-coding-inspired objective, called contrastive predictive coding, to learn useful abstract latent representations from unsupervised inputs \citep{oord2018representation}. The intuition behind contrastive predictive coding is that direct prediction in the data-space is often highly redundant, since it is often sufficient to model only short range temporal correlations in order to do well, and thus leads to models which overfit to minor details or flows than learning the true latent structure of the data. Moreover, if the objective is in the data-space, the model is penalized for mis-predicting small, irrelevant details, which can often lead it to devote modelling capacity to local noise instead of the global structure. Contrastive predictive coding instead argues to use predictive losses in the latent space of the model, so that the important prediction is of the future latent-state of the model itself, and not the actual observations. The original work showed that this approach worked well for learning informative representations of audio and visual datasets, while later work has demonstrated its benefits in video \citep{wen2018deep,elsayed2019reduced}. Interestingly, the standard Rao and Ballard predictive coding model implicitly implements this scheme as well, since apart from the lowest layer, all prediction errors are in the latent-space of the model and not the observation space. Effectively, predictive coding optimizes the sum of contrastive losses at every level of the hierarchy. It remains an open question whether such an objective would be beneficial for deep neural networks.

Finally, a small number of works have experimented with deep Rao and Ballard-style networks with additional sparsity regularisers \citep{chalasani2013deep}, which demonstrate a close link between predictive coding and solutions to sparse linear equation solvers and non-negative matrix factorization algorithms \citep{lee2001algorithms}. Overall, this is still a young area of research, with many open areas for further exploration. In general, the task of accurately translating the predictive coding paradigm into a modern deep learning framework, still remains open to new contributions, and additionally, it is still largely unknown the degree to which implementing recurrent and top-down connections (which comprise the majority of connections in the cortex \citep{kietzmann2019recurrence}) in artificial feedforward neural networks can improve performance and potentially lead to more biologically plausible, or robust machine learning solutions remains unknown. 

\section{Relationship to Other Algorithms}

\subsection{Predictive Coding and Backpropagation of error}

One of the major streams of technological advancement in the past decade has been in machine learning, with extremely large and deep neural networks, often containing millions or billions of parameters, to reach extremely high levels of performance on a wide range of extremely challenging tasks such as machine translation and text generation \citep{brown2020language,radford2019language}, Go \citep{silver2017mastering}, atari \citep{schrittwieser2019mastering}, as well as image \citep{child2020very} and audio generation \citep{oord2016wavenet,dhariwal2020jukebox} and object detection \citep{krizhevsky2012imagenet}. Core to all these successes is the backpropagation of error (backprop) algorithm \citep{werbos1982applications,linnainmaa1970representation}, which is used to train all such networks. Backprop is fundamentally a credit assignment algorithm which correctly computes the derivative of each parameter (often interpreted as a synaptic weight from a neuroscientific perspective) with respect to a global, often distant loss. Given these derivatives, or credit assignments, each parameter can then be adjusted independently in a way which will minimize the global loss. In such a way the network learns a set of weights which allows it to successfully solve the task. Crucially, backprop can compute gradients for effectively any computation, so long as it is differentiable. All that is needed, then, is for the operator to define a loss function, and the forward computation of the model, and then backprop can compute the gradients of each parameter in the model with respect to the loss. 

Due to the immense successes of backprop in training deep artificial neural networks, a natural question is whether the brain -- which faces an extremely similar credit assignment problem -- might potentially be using an algorithm like backprop to update its own synaptic weights. Unfortunately, backprop is generally not considered biologically plausible \citep{crick1989recent}, rendering a direct implementation of the algorithm in neural circuitry unlikely. However, in recent years a large amount of work has explored various biologically plausible approximations to, or alternatives to backprop, which could in theory be implemented in the brain \citep{bengio2015early,bengio2017stdp,sacramento2018dendritic,akrout2019deep,lillicrap2016random,whittington2017approximation,whittington2019theories,millidge2020relaxing,millidge2020activation,millidge2020predictive,millidge2020investigating,scellier2017equilibrium,scellier2018generalization,lee2015difference,amit2019deep}. This work reignites the idea that, in fact, biological brains could be implementing backpropagation for learning, or something very close to it. If this were the case, then this would provide an extremely important insight, allowing us to mechanistically understand many aspects of brain function and dysfunction, as well as allowing the immediate portability of results from machine learning, where experience with extremely large and deep backprop-trained neural networks exists, directly to neuroscience. It would also take a large step towards answering deep questions about the the nature of (human/biological) learning, and would make a substantial contribution towards our understanding of the prospects for the development of artificial general intelligence within the current machine learning paradigm.

From the perspective of this review, we focus on recent work investigating the links between predictive coding and the backpropagation of error algorithm. Specifically, it has been shown that, under certain conditions, the error neurons in predictive coding networks converge to the gradients computed by backprop, and that if this convergence has occurred, then the weight updates of predictive coding are identical to those of backprop. This was shown first in multi-layer perceptron models by \citet{whittington2017approximation} and then for arbitrary computational graphs, including large-scale machine learning models by \citet{millidge2020predictive} under the fixed prediction assumption. Similarly \citet{song2020can} has shown that if the network is initialized to its feedforward pass values, the first iteration of predictive coding is precisely equal to backprop. Since predictive coding is largely biologically plausible, and has many potentially plausible process theories, this close link between the theories provides a potential route to the development of a biologically plausible alternative to backprop, which may be implemented in the brain. Additionally, since predictive coding can be derived as a variational inference algorithm, it also provides a close and fascinating link between backpropagation of error and variational inference. 

Here we demonstrate the relationship between backprop and predictive coding on arbitrary computation graphs. A computation graph is the fundamental object which is operated on by backprop, and it is simply a graph of the computational operations which are undertaken during the computation of a function. For instance, when the function is a standard multi-layer perceptron (MLP), the computation graph is a series of elementwise nonlinearities and multiplication by the synaptic weights $y = f_1(\theta_1 f_2(\theta_2 f_3(\theta_3 \dots )))$.

Formally, a computation graph $\mathcal{G}(\mathbb{E},\mathcal{V})$ is a graph of vertices $\mathcal{V}$ and edges $\mathbb{E}$ where the vertices represent intermediate computation products -- for instance the activations at each layer in a MLP model -- and the edges represent differentiable functions. A vertex $v_i$ may have many children, denoted $\mathcal{C}(v_i)$ and many parents, denotes $\mathcal{P}(v_i)$. A vertex is a child of another if the value of the parent vertex is used to compute value of the child vertex, using the function denoted by the edge between them. For backpropagation, we only consider computation graphs that are directed acyclic graphs (DAGS) which ensure that it is impossible to be stuck in a cycle forever by simply going from children to parents. DAG computation graphs are highly general and can represent essentially any function that can be computed in finite time. Given an output vertex $v_{out}$ and a loss function $L = f(v_{out})$, then backpropagation can be performed upon a computation graph. Backpropagation is an extremely straightforward algorithm which simply uses the chain rule of multivariable calculus to recursively compute the derivatives of children nodes from the derivatives of their parents.
\begin{align*}
    \label{backprop_equation}    
    \frac{\partial L}{\partial v_i} = \sum_{v_j \in \mathcal{P}(v_i)} \frac{\partial L}{\partial v_j}\frac{\partial v_j}{\partial v_i} \numberthis
\end{align*}
By starting with the output gradient $\frac{\partial L}{\partial v_{out}}$, and if all the gradients $\frac{\partial v_j}{\partial v_i}$, which are the gradients of the edge functions, are known, then the derivative of every vertex with respect to the loss can be recursively computed. Then, once the gradients with respect to the vertices are known, the gradients with respect to the weights can be straightforwardly computed as,
\begin{align*}
\frac{\partial L}{\partial \theta_i} = \frac{\partial L}{\partial v_i}\frac{\partial v_i}{\partial \theta_i} \numberthis
\end{align*}
Predictive coding can also be straightforwardly extended to arbitrary computation graphs. To do so, we simply augment the standard computation graphs with an additional error units $\epsilon_i$ for each vertex. Formally, the augmented graph becomes $\tilde{\mathcal{G}}=(\mathcal{V}, \mathbb{E},\mathcal{E})$ where $\mathcal{E}$ is the set of all error neurons. We then adapt the core predictive coding dynamics equations from a hierarchy of layers to arbitrary graphs,
\begin{align*}
    \frac{dv_i}{dt} &= -\frac{\partial \mathcal{F}}{\partial v_i} = \epsilon_i - \sum_{j \in \mathcal{C}(v_i)} \epsilon_j \frac{\partial \hat{v}_j}{\partial v_i} \numberthis \\
\end{align*}
\begin{align*}
    \frac{d\theta_i}{dt} &= -\frac{\partial \mathcal{F}}{\partial \theta_i} = \epsilon_i \frac{\partial \hat{v}_i}{\partial \theta_i}  \numberthis
\end{align*}
where we have assumed that all precisions are equal to the identity $\Sigma_i^{-1} = \mathbb{I}$, and thus can be ignored. The dynamics of the parameters of the vertices $v_i$ and edge functions $\theta$ such that $\hat{v_i} = f(\mathcal{P}(v_i); \theta)$ can be derived as a gradient descent on $\mathcal{F}$, where $\mathcal{F}$ is the sum of prediction errors of every node in the graph. Importantly these dynamics still require only information (the current vertex value, prediction error, and prediction errors of child vertices) locally available at the vertex. Here we must apply the fixed prediction assumption and postulate that the prediction $\hat{v_i}$ remain fixed throughout the optimization of the $v_is$. Given this assumption, we can see that, at the equilibrium of the dynamics of the $v_is$, the prediction errors equals the sum of the prediction errors of the child vertices multiplied by the gradient of the prediction with respect to the activation,
\begin{align*}
    \label{PC_backprop_update}
    \frac{dv_i}{dt} = 0 \implies \epsilon^*_i = \sum_{j \in \mathcal{C}(v_i)} \epsilon_j^* \frac{\partial \hat{v}_j}{\partial v_i} \numberthis
\end{align*}
Importantly, this recursion is identical to that of the backpropagation of error algorithm (Equation \ref{backprop_equation}), which thus implies that if the output prediction error is the same as the gradient of the loss $\epsilon_{out} = \frac{\partial L}{\partial v_{out}}$, then throughout the entire computational graph, the fixed point of the prediction errors is the gradients computed by backprop, and thus by running the dynamics of Equation \ref{PC_backprop_update} to convergence, the backprop gradient can be computed. Finally, we see by inspection that the update rule for the weights in predictive coding and backpropagation are the same if the prediction error is equal to the gradient with respect to the activation, such that if the predictive coding network is allowed to converge, and then the weights are updated, then the procedure is equivalent to a single update of backpropagation. This means that, under the fixed prediction assumption, the dynamical iterative inference procedure of predictive coding can be interpreted as a convergence to the backpropagated gradients in an artificial neural network. While this equivalence is easy to see for an artificial neural network, there remain several issues of biological plausibility when applying this approach to neural circuitry. These are discussed in Appendix C.

Another set of results from \citet{song2020can} and \citep{salvatori2021predictive} demonstrate that the update in predictive coding is equal to backpropagation for the first update step even without this fixed prediction assumption. This is because for the first steps, all the prediction errors are 0 except at the output, where the prediction error is exactly the gradient of the loss function, and thus the last step will be identical to backpropagation in any case. Then, at each new step up to the N number of layers, the same process repeats with a new layer where the prediction error has been initialized to 0, and thus the first update is identical to the backpropagation one.

Intuitively, we can think of this as a predictive coding network in which all the error is initially focused at the output of the network, where the loss is, and then through the dynamical minimization of prediction errors at multiple layers, this error is slowly spread out through the network in parallel until, at the optimum distribution, the error at each vertex is precisely the credit that it should be assigned to causing the error in the first place. Importantly, unlike backprop, the dynamics of predictive coding are local, only requiring the prediction errors from the current vertex and the child vertices.

Finally, this correspondence implies a link between variational inference and backprop. Specifically, that backprop arises as a variational inference algorithm which attempts to infer (under Gaussian assumptions) the value of each vertex in the computational graph, given a known start and end node. Backprop can then be seen as the optimal solution to the `backwards inference' problem of inferring the values of the nodes of the graphs given a Gaussian prior centered at the activations provided by the feedforward pass. An additional further note is that the Gaussian assumptions of predictive coding gives rise to precision parameters $\Sigma^{-1}$ which are ignored in this analysis showing convergence to backpropagation. If these precisions are added back in, we see that it is possible to derive an `uncertainty-aware' extension to standard backprop,  which can adaptively regulate the importance of gradients throughout the computational graph depending on their intrinsic variance. This allows us to directly understand the assumption, implicit within backprop, that all nodes of the computational graph, and the data, are equally certain or uncertain, and that they are i.i.d distributed. The use of precision parameters, thus, may allow for the mathematically principled extension of backprop into situations where this implicit assumption does not hold. Exploring the close connections between backprop and inference is an exciting avenue for future work, which has recently been unlocked by the discovery of these important correspondences.

\subsection{Linear Predictive Coding and Kalman Filtering}
Here, we demonstrate how predictive coding in the linear regime corresponds to the celebrated Kalman Filter, a ubiquitous algorithm for optimal perception and filtering with linear dynamics \citep{kalman1960new}. The Kalman filter, due to its simplicity, accuracy, and robustness to small violations in its assumptions, is widely used to perform perceptual inference and filtering across a wide range of disciplines \citep{welch1995introduction} and is especially prevalent in engineering and control theory. Kalman filtering assumes a linear state-space model of the world defined as
\begin{align*}
    \mu_{t+1} &= A\mu_t + Bu_t + \omega \\ 
    o_t &= C\mu_t \numberthis
\end{align*}
where $\mu$ is the latent state, $u$ is some control or action input (for pure perception without any action on the world these terms can be ignored -- i.e. $B=0$ or $u = 0$), $\omega$ is a vector of white Gaussian noise, and $o$ is a vector of observations. These quantities are related through linear maps, parametrised as matrices $A$, $B$, and $C$.

This state-space model can be written as a Gaussian generative model.
\begin{align*}
p(o_{t+1},\mu_{t+1}, \mu_t) &= p(o_{t+1} | \mu_{t+1})p(\mu_{t+1} | \mu_t) \\
&= \mathcal{N}(o_{t+1}; C\mu_{t+1}, \Sigma_2)\mathcal{N}(\mu_{t+1}; A\mu_t + Bu_t, \Sigma_1) \numberthis
\end{align*}

Kalman filtering proceeds in two steps.  First the state is `projected forwards' using the dynamics model, or prior $p(\mu_{t+1} | \mu_t)$. Then these estimates are 'corrected' by new sensory data by inverting the likelihood mapping $p(o_{t+1} | \mu_{t+1})$. The Kalman filtering equations are as follows:

\textbf{Projection:}
\begin{align*}
    & \hat{\mu}_{t+1} = A\mu_t + Bu_t  &\\ 
    & \hat{\Sigma}_{t+1} = A\Sigma_t A^T + \Sigma_t \numberthis
\end{align*}
\textbf{Correction:}
\begin{align*}
    & \mu_{t+1} = \hat{\mu}_{t+1} + K(o_{t+1} - C\hat{\mu}_{t+1}) \numberthis
\end{align*}
\begin{align*}
    & \Sigma_{t+1} = (I - K)\hat{\Sigma}_{t+1} \numberthis
\end{align*}
\begin{align*}
    & K = \hat{\Sigma}_{t+1}C^T[C\hat{\Sigma}_{t+1}C^T + \Sigma_2]^{-1} \numberthis
\end{align*}
where $\hat{\mu}$ and $\hat{\Sigma}$ are the predicted values of $\mu$ and $\Sigma$ before new data are observed, and $K$ is the Kalman Gain matrix which plays a crucial role in the Kalman Filter update rules. The derivation of these update rules is relatively involved. For a concise derivation, see Appendix A of \citet{millidge2021neural} The estimated $\mu_{t+1}$ and $\Sigma_{t+1}$ found by the Kalman filter are optimal in the sense that they minimize the squared residuals $(\mu_{t+1} - \mu_{t+1}^*)^2$ and $(\Sigma_{t+1} - \Sigma_{t+1})^2$ where $\mu_{t+1}^*$ and $\Sigma_{t+1}^*$ are the `true' values, given that the assumptions of the Kalman filter (linear dynamics and Gaussian noise) are met. Kalman filtering can also be interpreted as finding the optimum of the maximum-a-posteriori estimation problem
\begin{align*}
    \mu^* &= \underset{\mu}{argmax} \, \, p(\mu | o) \\
    &\propto \underset{\mu}{argmax} \, \, p(o, \mu) \numberthis
\end{align*}
Since the generative model is assumed to be linear, this optimization problem becomes convex and can be solved analytically, giving the Kalman Filter solution.

Importantly, the optimization problem can also be solved iteratively using gradient descent. First, we write out the maximization problem explicitly,
\begin{align*}
    & \underset{\mu_{t+1}}{argmax} \, \,  p(\mu_{t+1} | o_{t+1}, \mu_t) \propto \underset{\mu_{t+1}}{argmax} \, \, p(o_{t+1} |\mu_{t+1})p(\mu_t+1 | \mu_t)  &\\
    &= \underset{\mu_{t+1}}{argmax} \, \, N(y=o_{t+1};C\mu_{t+1}, \Sigma_2)N(\mu_{t+1}; A\mu_t + Bu_t, \Sigma_1) \\
     &= \underset{\mu_{t+1}}{argmin} \, \, -o_{t+1}^T\Sigma_2 o_{t+1} + 2\mu_{t+1}^TC^T\Sigma_2 C\mu_{t+1} - \mu_{t+1}^TC^T\Sigma_2 y +\mu_{t+1}^T\Sigma_1\mu_{t+1} -2\mu_{t+1}\Sigma_1 A\mu_t -2\mu_{t+1}\Sigma_1 Bu_t \numberthis
\end{align*}

Given this loss function, to derive the dynamics with respect to $\mu_{t+1}$ we can take derivatives of the loss, which results in a familiar update rule.
\begin{align*}
    \frac{dL}{d\mu_{t+1}} &= 2C^T\Sigma_2 y - (C^T \Sigma_2 C + C^T\Sigma_2^T C)\mu_{t+1} + (\Sigma_1 + \Sigma_1^T)\mu_{t+1} - 2\Sigma_1 A\mu_t - 2\Sigma_1 Bu_t & \\
    &= -C^T \Sigma_2[o_{t+1} - C\mu_{t+1}] + \Sigma_1[\mu_{t+1} - A\mu_t - B\mu_t] \\
    &= -C^T \Sigma_2 \epsilon_z +  \Sigma_1 \epsilon_x \numberthis
\end{align*}
where $\epsilon_o = o_{t+1} - C\mu_{t+1}$ and $\epsilon_x = \mu_{t+1} - A\mu_t - Bu_t$.

These derivations show that predictive coding in the linear regime is an iterative form of the same optimization problem that the Kalman filter performs analytically. In effect, we see that predictive coding reduces to Kalman Filtering in the linear case, as the optimization procedure, due to the convexity of the underlying loss function, will converge rapidly and robustly. This insight not only connects predictive coding to well established optimal filtering schemes, and also showcases the more general nature of predictive coding since. Unlike Kalman filtering, predictive coding makes no assumptions of linearity and is able to perform effective perception and filtering even when the likelihood and dynamics models are potentially highly nonlinear, which is essential for systems (like the brain) operating in highly complex nonlinear worlds. 

Interestingly, by viewing the Kalman filtering problem in this probabilistic lens, it also allows us to straightforwardly derive an EM algorithm to \emph{learn} the matrices specifying the likelihood and dynamics models in the Kalman Filter. By taking gradients of the loss function with respect to these matrices, we obtain (see Appendix D for detailed derivations both of these update equations and the derivation of the Kalman Filter update rules):

\begin{align*}
    \frac{dL}{dA} &= -\Sigma_1 \epsilon_x \mu_t^T \numberthis
\end{align*}
\begin{align*}
    \frac{dL}{dB} 
    &= - \Sigma_1 \epsilon_x u_t^T \numberthis
\end{align*}
\begin{align*}
    \frac{dL}{dC} 
    &= -\Sigma_2 \epsilon_o \mu_{t+1}^T \numberthis
\end{align*}
We see that these dynamics take the form of simple Hebbian updates between the relevant prediction error and the estimated state, which could in theory be relatively straightforwardly implemented in neural circuitry.

\subsection{Predictive Coding, Normalization, and Normalizing Flows}

Computational accounts of brain function have stressed the importance of normalization, especially at the lower levels in perceptual hierarchies \citep{carandini2012normalization}. Sensory stimuli are almost always highly redundant in both space (close sensory regions are generally similar) and in time (the world typically changes smoothly, so that large parts of the sensory input are relatively constant over short timescales). A substantial portion of the brain's low-level circuitry (i.e. at the sensory epithelia and first few stages of processing) can be understood as accounting for and removing these redundancies. For instance, amacrine and ganglion cells in the retina are instrumental in creating the well-known centre-surround structure of receptive fields in the earliest levels of visual processing. Centre-surround receptive fields can be interpreted as the outcome of spatial normalization -- obtained by subtracting an expected uniform spatial background from the input. The brain also performs significant temporal normalization by subtracting away input that remains constant over time. A noteworthy example of this comes from the phenomenon of retinal stabilization \citep{riggs1953disappearance,gerrits1970artificial,ditchburn1955eye} whereby if you stare at a pattern for sufficiently long it will fade from conscious perception. Moreover, if a visual stimulus is experimentally held at a fixed location on the retina, fading is extremely rapid, often in less than a second \citep{coppola1996extraordinarily}, and can be explained straightforwardly by predictive coding of temporal sequences \citep{millidge2019fixational}.

The ubiquity of normalization in early sensory processing speaks to the utility and applicability of predictive coding models, since prediction and normalization are inseparable. Any normalization requires an implicit prediction, albeit potentially a crude one. Indeed, the earliest neurobiological predictive coding models \citep{srinivasan1982predictive} were deployed to model the normalization abilities of retinal ganglion cells. Spatial or temporal normalization is straightforward to implement in a predictive coding scheme. Consider the standard mean normalization whitening filter (Equation \ref{whitener_equation}) or the single step autoregressive whitener (Equation \ref{temporal_whitener})
\begin{align*}
    \label{whitener_equation}
    \hat{x} &= \frac{x - \mathbb{E}[x]}{\mathbb{V}[x]}  \numberthis
\end{align*}
\begin{align*}
    \label{temporal_whitener}
    \hat{x}_{t+1} &= \frac{x_{t+1} - x_t}{\mathbb{V}[x]} \numberthis
\end{align*}
Where $\mathbb{E}[]$ and $\mathbb{V}[]$ denote the expectation values and variances respectively. In these cases we see that the whitened prediction $\hat{x}$ is equivalent to the prediction errors in the predictive coding framework, with straightforward predictions of $\mathbb{E}[x]$ for the standard whitening filter and $x_t$ -- the value at the last timestep -- for the temporal whitener. The fact that precision-weighted prediction errors are the primary quantity transmitted `up' the hierarchy now becomes intuitive -- the prediction errors are effectively the inputs after normalization while the precision weighting instantiates that normalization.

Recently, this deep link between predictive coding and normalization has been further extended by \citet{marino2020predictive} by situating predictive coding within the broader class of \emph{normalizing flows}. Normalizing flows provide a general recipe for building or representing a complex distribution from a simple and tractable one, and have been recently applied with much success in challenging machine learning tasks \citep{rezende2015variational,papamakarios2019normalizing}. The central observation at the centre of normalizing flows is the 'change of variables' formula. Suppose we have two values $o$ and $\mu$ related by an invertible and differentiable transformation $f$ such that $o = f(\mu)$, then if instead we maintain distributions $p(o)$ and $p(\mu)$ then, by the change of variables formula, we know that,
\begin{align*}
    p(\mu) = p(o)|\frac{\partial f^{-1}}{\partial \mu}| \numberthis
\end{align*}
where $|\cdot |$ denotes the determinant and $\frac{\partial f^{-1}}{\partial \mu}$ denotes the Jacobian matrix of the inverse of the transform $f$. Crucially, if we can compute this Jacobian inverse determinant, then we can sample from or compute probabilities of $\mu$ given probabilities of $o$ and vice-versa. This allows us to take a simple distribution such as a Gaussian, which we know how to sample from and compute probabilities with, and then map it through a complex transformation to obtain a complex distribution while retaining the mathematical tractability of the simple distribution. A normalizing flow model can be constructed from any base distribution $p(\mu)$ as long as the transformations $f$ are invertible and differentiable. For instance, the transformation in Equation \ref{whitener_equation} is clearly invertible and differentiable (indeed it is affine) and thus constitutes a normalizing flow. The key insight in \citep{marino2020predictive}, is that most such normalization schemes can be considered to be simple normalizing flow models, which allows a rich theoretical analysis as well as their unification under a simple framework. Indeed, even the complex hierarchical predictive coding models developed later constitute normalizing flows as long as the functions $f(\theta \mu)$ are invertible and differentiable. The invertibility condition is hard to maintain, however, since it ultimately requires that the synaptic weight matrix $\theta$ is full-rank and square, which implies that the dimensionality of an invertible predictive coding network remains the same at all levels. This rules out many architectures which have different widths at each layer, such as standard autoencoder models which possess an information bottleneck \citep{tishby2000information}. Nevertheless, this close link between predictive coding architectures and normalizing flows allows us to immediately draw parallels between and understand the function of predictive coding networks as progressively normalizing and then mapping sensory stimuli into spaces where they can be more easily classified or form useful representations.

\subsection{Predictive Coding as Biased Competition}

The biased competition theory of cortical information processing is another highly influential theory in visual neuroscience. It proposes that cortical representations are honed through a process of lateral inhibitory competition between neighbouring neurons in a layer, where this competition is biased by top-down excitatory feedback which preferentially excites certain neurons which best match with the top-down attentional preferences, and this feedback enhances their activity, helping them to inhibit the activity of their neighbours and thus ultimately win the competition \citep{desimone1995neural,desimone1998visual,reynolds1999competitive}. Biased competition theory relies heavily on the notions of lateral feedback in the brain, and is supported by much empirical neurophysiological evidence of the importance of this kind of lateral feedback which is largely ignored in the standard predictive coding theory.

It initially appears that the predictive coding and biased competition are incompatible with one another, and that they make rival predictions. Perhaps the most obvious discrepancy is that predictive coding predicts inhibitory top-down feedback while biased competition requires top-down feedback to be excitatory, which is more in line with neurophysiological evidence throughout the cortex \footnote{Although this analysis discounts potential intralaminar circuitry -- especially interneurons -- which could potentially flip the sign of the effective connectivity.}. However, \citet{spratling2008reconciling} showed that in the linear regime these two theories are actually mathematically identical. The superfiial contrast between biased competition and predictive coding theories can be resolved by noting that they imply the same mathematical structure, which can be realized in multiple ways, with different neural circuits which have different patterns of excitation and inhibition. This unification shows how neurophysiological process theories should not necessarily be literal translations of the mathematics into neural groupings, and that moreover disproving one element of a process theory -- i.e. predictive coding requires top-down inhibitory feedback -- does not necessarily show that the theory is wrong, just that the process theory could be rewritten under a different rearrangement of the mathematical structure. It is important, therefore, to not treat the process theory, nor the standard form of the mathematics too literally.


The unification between predictive coding and biased competition is remarkably simple, requiring only a few straightforward mathematical manipulations. To show this, we first introduce a standard biased competition model \citep{harpur1994experiments}, rewritten in our standard notation to make the equivalence clear. Consider a layer of neurons $\mu$ with inputs $o$. This layer also receives top-down excitatory input $\theta_2 \bar{\mu}$ which is mapped through the top-down weights $\theta$. The neurons $\mu$ then inhibit their own inputs $o$ by mapping downwards through an inhibitory weight matrix $\theta_1$. Writing down these equations, we obtain
\begin{align*}
    \epsilon_o &= o - \theta_1 \mu \numberthis
\end{align*}
\begin{align*}
    \mu_{t+1} &= \alpha \mu_t + \beta \theta_1^T \epsilon_o + \gamma \theta_2 \bar{\mu}_t \numberthis
\end{align*}
Where $\alpha$,$\beta$,$\gamma$ are scalar constants which weight the importance of the different terms to the update. Similarly, we can write down the predictive coding update from Equation \ref{PC_update_rules}, but this time discretising the continuous time differential equation using the Euler integration scheme such that $\frac{dx}{dt} \rightarrow x_{t+1} = x_t + \eta \frac{dx}{dt}$ where $\eta$ is a scalar learning rate. By writing Equation \ref{PC_update_rules} as a discrete-time update, and setting all precisions to the identity $\Sigma^{-1} = I$, we obtain,
\begin{align*}
    \mu_{t+1} &= \mu_t + \eta \big( \epsilon_o \theta^T - \epsilon_x \big) \\
    &= \mu_t + \eta \epsilon_o \theta_1^T - \eta (\mu_t - \theta_2 \bar{\mu}_t) \\
    &= (1 - \eta) \mu_t + \eta \epsilon_o \theta_1^T + \eta \theta_2 \bar{\mu}_t \numberthis
\end{align*}

Thus, by equating coefficients $\alpha = 1- \eta$, $\beta = \eta$,$\gamma = \eta$, we see that the predictive coding and biased competition models are mathematically identical.

\subsection{Predictive Coding and Active Inference}
Predictive coding can also be extended to include action, allowing for predictive coding agents to undertake adaptive actions without any major change to their fundamental algorithms. The key insight here is to note that there are, in fact, two ways of minimizing prediction errors. The first is to update predictions to match sensory data, which corresponds to classical perception. The second is to take actions in the world to force the incoming sensory data to match the predictions \citep{friston2009reinforcement,friston2010action,clark2015surfing}. While seeming somewhat convoluted, this intuition for action can be formalized neatly and precisely within the mathematical apparatus of predictive coding. Specifically, one simply assumes that observations are a function of actions, which renders the free energy functional implicitly action-dependent. Then, one can simply minimize $\mathcal{F}$ with respect to action directly. This unification of perception and action within a predictive coding framework is usually referred to as `active inference' in the literature, since the core idea is that an agent can use the same complex generative model it uses for inference to enable it to take effective, adaptive actions. This kind of active inference using predictive coding was first proposed in \citet{friston2009reinforcement}, and has been developed in a number further works from both the theoretical \citep{pezzulo2016active,pezzulo2018hierarchical,friston2010action, adams2013predictions} and experimental \citep{baltieri2018probabilistic,baltieri2019pid,tschantz2020reinforcement,millidge2020relationship} sides. 

The basic approach to including action within the predictive coding framework is to simply minimize the variational free energy with respect to action. Although the free-energy is not explicitly a function of action, it can be made so implicitly by noticing the dependence of sensory observations on action. We can make this implicit dependence explicit using the chain-rule of calculus,

\begin{align*}
    \frac{da}{dt} = -\frac{\partial \mathcal{F}(o(a), \mu)}{\partial a} &= -\frac{\partial \mathcal{F}(o(a), \mu)}{\partial o(a)}\frac{\partial o(a)}{\partial a} \\
    &= -\frac{\partial o(a)}{\partial a} \Sigma^{-1}_o \epsilon_o \numberthis
\end{align*}

Where $\frac{\partial o(a)}{\partial a}$ is a `forward model' which makes explicit the dependence of the observation upon the action and must be provided or learnt by the algorithm, in addition to the standard generative model for perception. If the forward model is known, then the action selection rule for predictive coding is simply the forward model multiplied by the prediction error between the observation and the predicted observation. To obtain adaptive action, there also needs to be a notion of goals, desires or targets introduced into the inference procedure. These correspond to reward functions in reinforcement learning, or the concept of utility in economics. These goal inputs are must be provided exogenously to the inference procedure. Typically, these goals are encoded as set-points in the predictive coding scheme, which mathematically correspond to prior beliefs.



Under such a scheme, the prediction error $\epsilon_o = o - o^*$ simply becomes the difference between the current observation and the target or set-point $o^*$. However, this raises the question of where these set-points and targets come from and how they are computed. A generic answer to this question is that these set-points can be inherited from evolutionary or ontogenetic imperatives, or supplied by other neural circuits involved in goal-directed behaviour and planning. For present purposes, we can simply take them as exogenously given variables.

Neurobiologically, this scheme could be implemented through descending predictions in the motor cortex which would set priors or set-points for specific goals to be reached at different levels of hierarchical abstraction. Then the actual movements would arise simply through prediction-error minimization, where the prediction errors are simply the difference between the current state and the set-point. The function of the motor cortex, under this view, would be to simply transform the desired set-points from highly abstract representations potentially created in hippocampus, prefrontal cortex, or high-level sensory areas, into concrete motor-level setpoints required for actual control of low-level motor responses. The real work of calculating these set-points would be offloaded to the complex sensory generative models assumed present in the brain, while all motor computation would consist of simple `reflex arcs' minimizing prediction errors to a set-point \citep{friston2010action,pezzulo2018hierarchical}. The idea that the motor cortex only functions effectively as a conduit for descending motor prediction has been argued to explain the `agranularity' of the motor cortex \citep{shipp2013reflections} which is almost the same as cortical tissue but is missing the layer 4 `input layer'.

\subsubsection{Costs of action}

It is also straightforward to model potential costs of action. In biological organisms the key such cost is energetic. It takes valuable calories to move muscles and perform actions. Mathematically, such costs can be modelled within the predictive coding framework by explicitly including action within the generative model, as follows: $p(o,\mu, a) = p(o | \mu)p(\mu)p(a)= \mathcal{N}(o;\mu,\sigma_o)\mathcal{N}(\mu;\bar{\mu}, \sigma_\mu)\mathcal{N}(a;\bar{a},\sigma_a)$ where $\bar{a}$ represents a prior action expectation or desired action set-point, which is typically set at 0, so that every action performed by the organism is met with a quadratic penalty. The dynamics for action arise as a descent on the variational free energy, as before, with an additional action cost term,
\begin{align*}
    \frac{da}{dt} = -\frac{\partial \mathcal{F}(o(a), \mu,a)}{\partial a} &= -\frac{\partial \mathcal{F}(o(a), \mu,a)}{\partial o(a)}\frac{\partial o(a)}{\partial a} - \frac{\partial \mathcal{F}(o(a),\mu,a)}{\partial a}\\
    &= -\frac{\partial o(a)}{\partial a} \Sigma^{-1}_o \epsilon_o - \Sigma^{-1}_a \epsilon_a \numberthis
\end{align*}

where $\epsilon_a = a - \bar{a}$ represents the divergence between the action taken and the set-point, and $\Sigma^{-1}_a$ is the action precision, which is effectively a coefficient which determines how salient the costs of action are against the benefits.

\subsubsection{Active inference and PID control}

Action through predictive coding also shares strong links with classical control theory \citep{baltieri2020kalman}. Specifically, it has been shown that the well-known PID (`proportional-integral-derivative') control method \citep{johnson2005pid} is a special case of predictive coding with generalized coordinates \citep{baltieri2019pid}.

Like predictive coding, PID control simply optimizes a system towards a set-point, and is thus ideal for simple regulatory systems such as thermostats or motors. Action is determined by three terms -- a proportional term which minimizes the distance between the current location and the setpoint, an integral term which minimizes the integral of this error over time, and a derivative term which minimizes the derivative of the error. The combination of these three terms produces a highly robust and simple control system which can be straightforwardly applied, with some tuning, to control almost any simple regulatory process. The control law for PID control can be written as
\begin{align*}
    a(t) =\underbrace{k_p \epsilon_t}_{\text{Proportional}} + \underbrace{k_i \int_0^{\infty} dt \epsilon(t)}_{\text{Integral}} + \underbrace{k_d \frac{d\epsilon(t)}{dt}}_{\text{Derivative}} \numberthis
\end{align*}
where the error $\epsilon(t) = o(t)  - o^*(t)$ measures the difference from the setpoint $o^*(t)$ at a specific time-step, and $k_p$, $k_i$, $k_d$ are scalar coefficients which weight the importance of each of the three terms and can be adaptively tuned to achieve the desired performance characteristics of the controller. Intuitively, the proportional term minimizes the immediate instantanous error, and thus drives the system towards the set-point. The integral term makes the system robust to any continued step-change \footnote{A step-change is a term from control theory which means a constant un-modelled input entering the system. For instance suppose you are trying to maintain the speed of an object but have not accounted for atmospheric drag -- the drag term will be a unmodelled constant force acting on the object.} disturbance as the integral will eventually grow large enough to drive the system back towards the set-point, and the derivative term helps dampen oscillations and make the controller more robust in general. The combination of these three terms provides a highly robust form of control which can be broadly applied without needing precise knowledge of the dynamics of the system to be regulated. Because of these properties, the PID controller is widely used in a vast range of industrial and commercial applications \citep{johnson2005pid}. Importantly, it has been demonstrated \citet{baltieri2019pid} that PID control emerges as a special case of predictive coding in generalized coordinates with a linear (identity) generative model, an identity forward model \citep{baltieri2019pid}. This equivalence therefore will also clarify the implicit assumptions made in PID control and how to potentially extend PID control to explicitly handle more complex situations, or to incorporate additional knowledge of the dynamics that a control engineer may possess into the algorithm in a mathematically justified way.

To obtain equivalence to predictive coding, we utilize a linear (identity) generative model with three levels of dynamical orders $p(\tilde{o}, \tilde{x}) = p( o | x)p(o' | x')p(o'' | x'')p(x | x')p(x' | x'')p(x'')$. We can write out this generalized model explicitly as,
\begin{align*}
    o &= x + z \\
    o' &= x' + z' \\
    o'' &= x'' + z'' \\
    \dot{x} &= x' = x - \bar{\mu} + \omega \\
    \dot{x'} &= x'' = x' - \bar{\mu'} + \omega'  \\
    \dot{x''} &= x''' = x'' - \bar{\mu''} + \omega''  \numberthis
\end{align*}

where $\bar{\mu}$ is the desired set-point for $x$ at that dynamical order. We assume a standard variational delta-distribution posterior which factorizes across generalized orders $q(x, x',x''; \mu,\mu',\mu'') = \delta(x - \mu)\delta(x' - \mu')\delta(x'' - \mu'')$. Given this generative model and variational posterior, we can write out the variational free energy as,
\begin{align*}
    \mathcal{F} = \frac{1}{2}\Big[ &\sigma_z (o - \mu)^2 + \sigma_{z'}^{-1} (o' - \mu')^2 + \sigma_{z'}^{-1}(o'' - \mu'')^2 + \sigma_{\omega}^{-1}(\mu' - (\mu - \bar{\mu})^2 \\ &+ \sigma_{\omega'}^{-1}(\mu'' - (\mu' - \bar{\mu'})^2 + \sigma_{\omega''}^{-1}(\mu''' - (\mu'' - \bar{\mu''})^2 \Big] \numberthis
\end{align*}

By taking the derivatives of the free energy with respect to $\tilde{\mu} = \{\mu, \mu', \mu''\}$, and compressing the three equations into one generalized coordinate equation we can write and solve for the fixed-point of these equations as follows
\begin{align*}
    D\tilde{\mu} &= \tilde{\sigma}^{-1}_{\omega}(\tilde{\mu} - \tilde{\bar{\mu}}) \\
    &\rightarrow D\tilde{\mu} = 0 \implies \tilde{\mu} = \tilde{\bar{\mu}} \numberthis
\end{align*}

Thus at the equilibrium of the perceptual dynamics, we have simply that the latent state is equal to the desired latent state. If we assume that the dynamics converge quickly, we can write the dynamics with respect to action as,
\begin{align*}
    \frac{da}{dt} = -\frac{\partial \mathcal{F}}{\partial \tilde{o}}\frac{\partial \tilde{o}}{\partial a} &= -\sigma_{z}^{-1}(o - \bar{\mu})\frac{\partial o}{\partial a} - \sigma_{z'}^{-1}(o' - \bar{\mu}')\frac{\partial o'}{\partial a} - \sigma_{z''}^{-1}(o'' - \bar{\mu}'')\frac{\partial o''}{\partial a} \numberthis
\end{align*}
If we then assume that there is an identity mapping between action and changes of observation at each dynamical level -- $\frac{\partial o}{\partial a}=\frac{\partial o'}{\partial a}=\frac{\partial o''}{\partial a}=\mathbb{I}$, and that the desired states at each dynamical order are simply the desired state at the lowest -- $\bar{\mu}' = \bar{\mu}'' = \bar{\mu}$ then active inference through predictive coding reduces to the derivative form of PID control,
\begin{align*}
    \frac{da}{dt} = -\sigma_{z}^{-1}(o - \bar{\mu}) - \sigma_{z'}^{-1}(o' - \bar{\mu}) - \sigma_{z'}^{-1}(o'' - \bar{\mu}) \numberthis
\end{align*}

where the precisions $\sigma^{-1}_z$ can be identified with the weighting coefficients $k$ in the PID algorithm. Although the equivalence between predictive coding and PID requires many assumptions which appear unjustified, in reality it is simply making explicit the implicit assumptions already made by PID control, and therefore shows ways that predictive coding can be used to generalize PID control to improve performance in situations where more information about the desired dynamics or set-points are known by relaxing these constraints using more complex predictive coding models. For instance, predictive coding immediately shows how to correctly handle systems with non-identity, and even arbitrarily nonlinear relationships between the action and the quantity to be controlled, through the specification of the forward model -- i.e. by replacing the identity forward model assumed here with a control matrix $B$ or an arbitrary nonlinear function $f$. Moreover, by noticing that the weighting coefficients in PID control simply are precisions, predictive coding provides a principled and straightforward way to \emph{learn} these coefficients as a gradient descent on the variational free energy, which can improve performance compared to the ad-hoc methods often proposed for tuning in the PID control literature \citep{baltieri2019pid}.

Beyond PID control, active inference through predictive coding also shares close links with a number of other algorithms in classical control theory \citep{baltieri2020bayesian}, and also provides a unified Bayesian variational framework for extending and deeply understanding such algorithms, and these relationships are not yet fully worked out and are an active area of current research.

Active inference in a predictive coding paradigm may also potentially be useful for modelling the kinds of regulatory feedback loops which are used to control the internal homeostatic state or organisms. For instance, there is a well-studied multi-stage feedback loop in the human body which control glucose levels in the blood \citep{alon2019introduction}. In effect there is a two-level negative feedback loop first between the glucose levels in the blood and beta cells in the liver which helps counteract glucose spikes after eating a meal, and the second level uses blood glucose levels to control the rate of division of beta cells, thus preventing either runaway growth or runaway death of these cells. It is possible to interpret such a system in the language of interoceptive predictive coding as PID control, however it is at this point unclear what benefit this interpretation will have over the standard way of thinking about this explicitly in terms of classical control and cybernetic feedback processes. Nevertheless, the modelling of interoceptive, homeostatic and allostatic feedback processes within the body using predictive coding  has already been explored to some degree under the name of interoceptive inference \citep{seth2012interoceptive,seth2013extending,seth2013interoceptive}. The key advantage of predictive coding is that it offers a well-tested and mathematically principled framework in which to model these phenomena. These advantages may allow the modelling of considerably more complex interoceptive feedback loops than can be straightforwardly interpreted in terms of feedback control \citep{barrett2015interoceptive,pezzulo2018hierarchical,tschantz2021simulating}, and especially allows for the relatively straightforward construction of nested and hierarchical feedback control models.

However, it is important not to over-extend interpretations of action-oriented predictive coding. Fundamentally, all the predictive coding framework is doing is minimizing immediate local prediction errors from a set-point. The real hard work in complex control problems arises precisely when it is not possible to simply minimize divergence from the goal, but instead when complex interrelated sequences of actions are necessary to attain some goal. In such a paradigm, constructs such as value functions are required which can take into account these nonlinear and temporal dependencies \citep{sutton2018reinforcement}. These value functions can either be estimated directly through various iterative formulations of the Bellman optimality principle, such as temporal difference or Q learning, which forms the foundation of model-free reinforcement learning, or else can be estimated directly through simulated roll-outs using a world model, which forms the basis of explicit planning and model-based reinforcement learning. Predictive coding cannot tackle these problems alone -- instead it assumes their solution in the creation of the requisite set-points. In this sense, then action through predictive coding simply cannot explain the full complexity of motor computations in the brain, and instead focuses only on the periphery which is involved in correcting small deviations from a desired set-point which is itself computed using more complex algorithms. Indeed, this is suggestive of a similar scheme widely used in robotics, whereby complex planning or reinforcement learning algorithms determine the desired forces to send to the actuators, however the actuators themselves are equipped with onboard PID controllers, to ensure that the expected force is actually produced by the motors, and to correct for minor deviations \citep{johnson2005pid} occuring at the lower levels of translating motor commands into action. Predictive coding, then, would be a model of these inbuilt controllers at the periphery rather than the core action selection mechanisms in the brain.
 
\section{Discussion and Future Directions}

In recent years, work on predictive coding has substantially extended the initial theories of \citet{rao1999predictive} and \citet{friston2003learning}. Theoretical advances have been accompanied by considerable empirical testing.  As a result, predictive coding approaches to cognition and cortical function have gained substantial influence in neuroscience, as well as stimulating a number of philosophical debates \citep{williams2018predictive,hohwy2008predictive,clark2013whatever,clark2015surfing,cao2020new,seth2014cybernetic,seth2014cybernetic}. 

We have seen that predictive coding provides a mathematically principled method for perception and learning under Gaussian assumptions, that predictive coding schemes can solve object recognition and other challenging cognitive tasks, that they possess well-supported neurobiologically realistic process theories that in turn give rise to many concrete neuroscientific predictions which are only just beginning to be systematically tested \citep{walsh2020evaluating}. Predictive coding can account for a wide range of neurobiological and psychological effects such as end-stopping \citep{rao1999predictive}, repetition and expectation suppression \citep{auksztulewicz2016repetition}, and bistable perception \citep{hohwy2008predictive}, and has even been applied to help understand psychiatric disorders such as autism \citep{lawson2014aberrant} and schizophrenia \citep{sterzer2018predictive}.  As such predictive coding stands as perhaps the most developed and empirically supported general theory of cortical function. In this review, we have surveyed many of the interesting avenues of the theory and its applications sufficiently for an interested reader to form a detailed understanding of the major strands of work. In the remainder of this paper, we discuss and evaluate this existing work, and propose future areas of focus which are either not sufficiently well understood, or else seem necessary in order to develop predictive coding into a fully-fledged general theory of cortical function. 

While process theories of predictive coding are well-developed, as discussed at length in the microcircuits section, there remain several aspects of  cortical microcircuitry which have so far resisted a simple interpretation using this framework. These include the backwards superficial to superficial feedback connections, the upwards connectivity from  deep layers back to L4, as well as the substantial interactions that all cortical regions have with subcortical regions, which primarily take the form of a cortico-subcortical loops where cortical information is projected to subcortex via the deep layer 5, which then projects back to the primarily superficial layers of the cortex. Or, alternatively, from the subcortical view, it projects to the superficial layers which then are transmitted downwards to the deep layers, which provide the cortical response. Precisely what the functional purpose of such loops is, and what information is transmitted, remains largely mysterious in general and does not have any explanation within standard formulations of predictive coding. Additionally, the role of lateral inhibitory connectivity within laminar regions, which is implemented through a dizzying array of different types of interneurons, is largely not explained within the predictive coding paradigm. Much of this connectivity may be implementing homeostatic plasticity and other basic functions \citep{turrigiano1999homeostatic,watt2010homeostatic}, or else could be implementing some kind of layer-wise normalization -- which should theoretically be describable as predictive coding -- but in general the exact function of this connectivity at a computational and algorithmic level is unknown both within and outside of predictive coding, as is the interaction between top-down and lateral connectivity. As such, it may be that predictive coding might be `right' in some sense, but still missing core aspects of the computation that actually goes on in the cortex. 

Another question is that it is still largely undecided what the right kinds of generative models are to understand cortical computation? The current paradigm is strongly wedded to Gaussian assumptions under hierarchical static or dynamical generative models. The explicit prediction error minimization framework only arises under Gaussian assumptions. Discrete generative models have, in the broader active inference literature, been found to be able to reproduce many neurophysiological and behaviour phenomena \citep{friston2017active,friston2017process,friston2015active}. However, although the updates take the form of a reasonably biologically plausible message passing scheme \citep{parr2019neuronal}, they do not directly involve explicit prediction error minimization. An interesting question, then, remains to be determined which is the extent of predictive coding, if it exists, throughout the brain. For instance, it appears likely  that the lowest levels of the brain such as the retina and the LGN implement predictive-coding like schemes \citep{srinivasan1982predictive,huang2011predictive} and potentially it may explain a substantial amount of higher sensory processing. Similarly, in action, the robustness, effectiveness of PID control and its relationship to predictive coding may imply that the lowest levels of the motor system also implement a predictive-coding like architecture for action. However, it is also possible that the highest and most abstract levels of the brain may be primarily discrete, necessitating discrete generative models for their mathematical description which would entail a departure from predictive coding. Similarly, adaptive action in biological organisms requires long-term planning and maximization of future reward signals, which cannot readily be implemented within the predictive coding paradigm but instead likely relies on a complex set of machinery specialised for performing reinforcement learning \citep{sutton2018reinforcement}. 

Another important aspect of computation in the brain involves memory, both short term working memory which appears to be implemented by sustained firing in cortical regions, as well as long-term memory which may be mediated by interaction between the hippocampal-entorhinal system and the prefrontal cortex \citep{remondes2004role}, both of which appear difficult to understand and model within the predictive coding framework as presented here, although there has been some research in this direction \citep{fountas2020predictive}. Generalized coordinates do provide the system with a short-term temporal memory, which can implement computations such as local temporal decorrelations or the removal of redundant optical flow, but in practice do not provide a sufficiently long term memory for cortical working memory, let alone long-term memory in the brain \footnote{In theory generalized coordinates do suffice, since they are simply a Taylor expansion of the function around the present time-point which, when extended to infinite order, can precisely express the entire function through time. In practice, we cannot compute with infinite orders, and usually truncate the generalized order to about 4-6, which only provides a kind of short-term dynamical memory.}. 

An additional question arises from the fundamental architecture of the cortex. For instance, predictive coding as currently implemented deals primarily with fully-connected layers of neurons, which ignores all that we know about sparse cortical connectivity and especially the seemingly columnar architecture of the cortex. Understanding what these neuroanatomical features `mean' in a computational and algorithmic sense, as well as how they may potentially map to, or inspire novel machine learning architectures remains unknown. Luckily, the mathematical framework of predictive coding is not closely coupled to the specifics of the generative models themselves as all this complexity can be absorbed within the prediction function $f$, however such differences in architecture could have profound effects on the proposed neural microcircuitry to actually implement the algorithm.

Another area of the predictive coding framework which is ripe for further theoretical and empirical development is the notion of precision. While much of the literature, especially of an intuitive and philosophical bent, arguing that precision is important for coordinating global modulatory responses and adding considerable flexibility to the predictive coding paradigm, exactly how to translate these ideas into the formal mathematics, neural process theories, and computer simulations is unclear. For instance, understanding precision simply as the variances of the prediction errors does not appear to fulfill the kind of modulatory role that is required of it. Gaining a new understanding of how modulation of feedforward input in the brain is implemented, as well as experimenting with its importance in large-scale simulations of predictive coding networks with simultaneous learning has not yet been done. Finally, the implementation of precision in the brain remains largely mysterious. There are competing theories of precision being implemented through lateral connectivity \citep{friston2005theory}, feedback superficial-to-superficial connectivity \citep{shipp2016neural}, or primarily through subcortical processes and regions such as the pulvinar \citep{kanai2015cerebral}. However, to our knowledge, there has been no real explicit experimental tests of the differing predictions made by these theories.

Additionally, there remain substantial challenges in scaling up predictive coding networks to address challenges at the forefront of modern ML, including many empirical questions over the best way to scale up such networks, the degree to which modern machine learning architectures such as CNNs, resnets, and transformers can be straightforwardly integrated into the predictive coding paradigm and any potential benefits of doing so, as well as the impacts of top-down connectivity upon the performance of the algorithm. Finally, while there has been considerable progress in predictive coding networks setup for the machine learning paradigm of hierarchical static image classification, there has been considerably little work experimenting with potentially more brain-like paradigm such as spatial predictive coding between pixels in an image, or temporal predictive coding over sequences. Indeed, there has not yet even been any large scale study of the performance characteristics of generalized coordinates on spatio-temporal video data, so precisely how effective they are in such settings remains unknown.

The astounding progress in machine learning in unsupervised generative modelling has largely confirmed the contention of predictive coding that unsupervised predictive, or autoregressive objectives suffice, when applied to extremely large and deep models trained on vast datasets of unlabelled data (which the brain certainly has no shortage of via its sensory streams) can both learn extremely powerful and general world models which exhibit extreme flexibility and generalizability as well as learning highly robust and abstract representations. This has been exhibited in the extremely impressive natural language capabilities of unsupervised autoregressive transformers such as the GPT-x family of models \citep{radford2019language,brown2020language}, the highly visually realistic images that can be created both with very large generative-adversarial \citep{goodfellow2014generative} and variational autoencoding models \citep{kingma2013auto,child2020very}, as well as recent work showing the considerable power of unsupervised contrastive objectives to learn powerful representations in audio \citep{oord2016wavenet}, and text and images \citep{radford2021learning}. The question now remains, is there anything additional that predictive coding can bring to machine learning? Perhaps the key idea that predictive coding possesses, which is lacking in current machine learning, is a strong role for recurrent, top-down connectivity. In many ways, the success of purely feedforward machine learning architectures (which even in the case of transformers outperform previously recurrent architectures such as the LSTM in natural language tasks) has retroactively bolstered much of the `classical' view of visual perception as a pure feedforward pass through a set of hierarchically composable feature detectors. This approach is precisely the one taken by large scale machine learning models, and has been shown to be able to achieve human-level object recognition and detection in images \citep{krizhevsky2012imagenet}, as well as matching to a surprising degree the observed activations in biological brains \citep{yamins2014performance}. Nevertheless, it is possible that performance may be further improved, and the models themselves may be made more flexible by the introduction of a recurrent top-down processing stream. There are many theoretical advantages of recurrence \citep{van2020going,kietzmann2019recurrence} such as effectively greater depth, and the ability to flexible choose how much computation (recurrent cycles) to spend on a stimulus. Whether these advantages appear straightforwardly in recurrent predictive coding models remains to be conclusively tested empirically, although the recent work in deep predictive coding \citep{lotter2016deep,wen2018deep} is a welcome and important step in this direction. One reductive perspective is that feedback and recurrence in the brain is simply attempting to mimic extreme depth in a feedforward network, which has been demonstrated empirically to be vital for making maximally expressive and powerful models, often summarised as the `blessings of scale' \citep{kaplan2020scaling}. The visual system, for instance, in the brain is not that deep. It comprises only some 10-30 `layers' \citep{felleman1991distributed} while state of the art visual object recognition systems are considerably deeper. Top-down feedback connectivity and the recurrence it generates, could then simply be the brain's approach of mimicking extreme depth in a space-constrained way. It is also possible that there are considerable benefits of feedback processing beyond simply mimicking a deeper feedforward pass. Implicit weight-sharing in the brain across timesteps during recurrence implicitly encodes a time-translation invariance as an inductive bias of the network, and may aid the brain naturally handle relatively slowly changing sequences, which is fundamentally the `native' datatype the brain must handle in its sensory streams, while machine learning algorithms are currently primarily specialised on static data-types such as still images or text sequences presented instantaneously and simultaneously to a transformer. 

Another interesting idea which arises naturally from predictive coding and neuroscience which has been little explored in machine learning is the potential of heterarchical and specialised architectures. Contemporary machine learning architectures are typically comprised of a strictly hierarchical set of modules which implement a monolithic feedforward pass, while the brain is much more heterarchical. While there is some hierarchy, it is not a ladder. There are many parallel streams and multiple areas which `branch-off' the main hierarchy at various points, as well as all cortical areas receive inputs from subcortical areas which can substantially modulate their processing. We believe that in the future, as machine learning architectures continue to grow and become more capable, as well as handling more multimodal inputs with potentially different loss functions being minimized simultaneously, that we will witness a similar growth of heterarchy and multiple current architectures which are each specialised for a specific task being joined together into a heterarchical very large network, potentially with a global workspace architecture \citep{vanrullen2021deep}, which can replicate much more of the general functionality of the brain. 



A particularly salient direction of current and future research is the relationship between predictive coding and backprop, both for potentially establishing fundamental principles of biological learning, and for potentially extending already powerful backprop algorithms in machine learning to the cortex. The most important issue is to understand how far the `fixed-prediction' assumption can be relaxed while still maintaining learning performance, as well as how rapidly the predictive coding inner inference loop can be allowed to converge. If the connection between predictive coding and backprop could be made fully biologically plausible, this would provide a different view on the importance of feedback connections in the brain, and render our understanding of neural computation much more similar to contemporary machine learning than we currently believe. For instance, under this perspective, the feedback connections would not be implementing any deliberate kind of modulation or top-down contextual processing on their targets. Any such effects would be mere epiphenomena of the necessity to transmit gradient signals backwards to subserve credit assignment. Furthermore, if this were true, it would provide a precise specification of the signals sent through the feedback connections -- the required backwards gradients and prediction errors needed to compute Equation \ref{backprop_equation}. Another hypothesis is that the brain combines both types of feedback processing -- feedback signals for credit assignment along the feedforward path, and secondarily modulatory or context-sensitive feedback processing which enables the development of contextual phenomena such as extra-classical receptive field effects, as well as global processing modulation such as attention. There would then be two separate feedback pathways in the brain. Such a separation of pathways even has some empirical support in the laminar structure of the cortical region, where there are in fact two feedback pathways -- a superficial to superficial and a deep to superficial and deep pathway. Under this hypothesis, the superficial to superficial feedback pathway would be transmitting backwards the required signals for credit assignment while the deep feedback pathway would be transmitting more modulatory predictions as well as information required for global processing such as attention. Under such an architecture there remains a further problem of how to successfully learn and backpropagate through the modulatory feedback path, and it remains unclear how this could be done, or whether it is even necessary. Understanding the dynamics of such networks is currently extremely underexplored and would likely be an extremely important avenue for future work. 

Additionally, a serious potential issue with much current work, including predictive coding, both approximating backpropagation and being used as a model for the brain is that these models assume a fundamentally rate-coded substrate while in reality the brain uses spiking neural networks both to transmit information and to learn. It is not yet clear the degree to which spiking neural networks are used to simply approximate the behaviour of rate-coded ones, or whether the brain actually utilizes precise spike-timing to convey information. A-priori, given the strong evolutionary pressures towards space and energetic efficiency that the biological brains have evolved under, it seems likely that if there is a straightforward way to utilize such information, the brain will use it. As such it is not clear whether rate-coded theories can simply be straightforwardly ported over to more biophysically realistic spiking models or not. This is a very important area for future work since it is vital to understand the degree to which predictive coding models can function in a spiking environment, and to design bespoke predictive coding algorithms for spiking neural networks, which can be empirically investigated for their performance and robustness characteristics as well as their biological plausibility.

A further necessity in understanding predictive coding in the brain is a strong grasp on the importance of \emph{time}. The brain is enmeshed in a continuous time world and is exposed to (and actively shapes) a continually varying sensory stream. This contrasts dramatically with many ML setups in which a network is passively exposed to data without consideration of temporal extension or dependencies. Moreover, processing in predictive coding takes time, time in which the world itself can have changed. The brain simply cannot spend large numbers of cycles iteratively refining perceptions of some sensory datum which may then have rapidly changed again. Understanding the importance of continuous time, including the necessary lags induced by the brain's own sensory system as well as its computation time itself is vital. So is discovering and developing generative models flexible enough to handle learning and inference on continuously changing sensory streams, to develop abstract predictive representations both of relatively static invariants in the data, but also modelling continuously changing inputs. Generalized coordinates and dynamical models are an important step in this direction, and are mathematically designed to handle the case of colored, or correlated, noise (i.e., noise with a non-zero autocorrelation). However, generalized coordinates only provide a very short term dynamical model and memory of the system, which is potentially ideal for modelling local smooth temporal dependencies such as optical flow, but is perhaps not sufficient for modelling more abstract changes over time. Developing and scaling flexible and expressive generative models to handle these kinds of temporal dependencies in continuously varying inputs is an open challenge in both neuroscience as well as in machine learning.

Continuous time also changes the problem the brain faces in perception from one of pure inference to one of filtering \citep{jazwinski2007stochastic,stengel1986stochastic}. Namely, it must combine both sensory observations with prior knowledge of the current state to form its sensory percept, not just sensory information alone. Creating large-scale predictive coding networks, and the mathematical formalism behind them, that are specialised for perceptual filtering and not inference is also an important avenue for development of the theory. The fact that in the linear case, predictive coding as filtering possesses such a close relationship to Kalman filtering is perhaps encouraging in this direction. Finally, continuous time also makes the credit assignment and learning task much harder. It is not sufficient to simply backpropagate information through a static computational graph. Instead, activity now affects events in the future, and this means that credit assignment must take place not just through space, but through time as well. The machine learning solution to this problem -- backpropagation through time (BPTT) -- is wildly implausible in biological and physical systems, since it explicitly requires sending information (credit signals) \emph{backwards through time}. To implement such a system in the brain would require neurons to store a precise memory of every spike they emitted and input they received, as well as their weight-state at every point in time in some window, to be able to compute the required backwards messages \citep{lillicrap2020backpropagation}. It is possible to transform BPTT into an algorithm which only requires looking forward in time, by effectively storing a trace of previous activities and gradients, known as real-time recurrent learning (RTRL) \citep{williams1989experimental}, and there have been a number of proposals of more biologically plausible or approximate versions of RTRL which might be potentially implemented in the brain \citep{tallec2017unbiased}. Overall, although much progress has been made in recent years in designing biologically plausible credit assignment algorithms for static computational graphs using rate-coded neurons, the real challenge of designing biologically plausible credit assignment algorithms for hierarchical spiking neural networks which must assign credit throughout temporal sequences is still largely open, although there has been some important recent work in this direction \citep{bellec2020solution,zenke2018superspike}.

\section*{Acknowledgements}
We would like to thanks Alexander Tschantz, Conor Heins, and Rafal Bogacz for useful discussions about this manuscript and on predictive coding in general.

\bibliography{cites}
\section{Appendix A: Predictive Coding Under the Laplace Approximation}

In the main derivation of the variational free energy $\mathcal{F}$, we used the assumption that the variational density is a dirac delta function: $q(x | o; \mu) = \delta(x - \mu)$. However, the majority of derivations, including the original derivations in \citep{friston2005theory} instead applied the Laplace approximation to the variational distribution $q$. This approximation defines $q$ to be a Gaussian distribution with a variance which is a function of the mean $\mu$:  $q(x | o; \mu) = \mathcal{N}(x;\mu, \sigma(\mu))$. Notationally, it is important to distinguish between the generative model $\Sigma$, and the variational distribution $\sigma$. Here we use the lower-case $\sigma$ to denote the parameter of the variational distribution. The lower-case is not meant to imply it is necessarily a scalar variable. As we shall see, the optimal $\sigma$ will be become the inverse-Hessian of the free energy at the mode. 

Intuitively, this is because the curvature at the mode of a Gaussian distribution gives a good indication of the variance of the Gaussian, since a Gaussian with high curvature at the mode (i.e. the mean) will be highly peaked and thus have a small variance, while a Gaussian with low curvature will be broad, and thus have a large variance.  While our derivation using a dirac-delta approximation and the standard derivation using a Laplace approximation obviously differ, they ultimately arrive at the same expression for the variational free energy $\mathcal{F}$. This is because both approximations effectively remove the variational variances from consideration and only use the variational mean in practice. The only difference in the approximations arises from the constant entropy term in the free energy, which under the dirac delta assumption is 0 since the entropy of a dirac delta distribution is 0, while under the Laplace approximation this entropy term is nonzero but constant with respect to parameters being optimized.  Under the Laplace approximation, the variational coveriance has an analytical optimal form and thus does not need to be optimized, and plays no real role in the optimization process for the $\mu$s either. In the main text, we chose to present our derivation using dirac-deltas in the interests of simplicity, however here we will present the full Laplace-approximation derivation and demonstrate the equivalence between the two for our purposes.

To begin, we return to the standard energy function in multilayer case, this time under the assumption of the Laplace approximation.

\begin{align*}
    \mathcal{F} &= \sum_{i=}^L \underbrace{\mathbb{E}_{q(x | o;\mu)}\big[ \ln p( x_i | x_{i+1}) \big]}_{\text{Energy}} + \underbrace{\mathbb{E}_{q(x | o;\mu)} \big[ \ln q(x | o ;\mu) \big]}_{\text{Entropy}} \\
    &= \sum_{i=}^L \mathbb{E}_{\mathcal{N}(x;\mu, \sigma(\mu))}\big[ \ln \mathcal{N}(x_i;f(\mu_{i+1}, \Sigma(\mu))) \big] - \ln 2 \pi \sigma_i \numberthis
\end{align*}
Where we have used the analytical result that the entropy of a Gaussian distribution $\mathbb{H}[\mathcal{N}] = \ln 2 \pi \sigma$.
Then, we apply a Taylor expansion around $x_i = \mu_i$ to each element in the sum,
\begin{align*}
    \mathcal{F} &\propto \sum_{i=}^L \mathbb{E}_q \big[ \ln p(\mu_i | \mu_{i+1}) \big] + \mathbb{E}_q \big[\frac{\partial \ln p(x_i | x{i+1})}{\partial x_i}(x_i - \mu_i) \big] + \mathbb{E}_q \big[\frac{\partial^2 \ln p(x_i | x{i+1})}{\partial x_i^2}(x_i - \mu_i)^2 \big]  - \ln 2 \pi \sigma_i\\
    &= \sum_{i=0}^L \ln p(\mu_i | \mu_{i+1})  +  \frac{\partial^2 \ln p(x_i | x{i+1})}{\partial x_i^2}\sigma_i - \ln 2 \pi \sigma_i \numberthis
\end{align*}
Where in the second line, we have used the fact that $\mathbb{E}_q \big[x_i - \mu_i \big] = \mathbb{E}_q \big[x_i] - \mu_i = \mu_i - \mu_i = 0$ and that $\mathbb{E}_q \big[ (x_i - \mu_i)^2] = \Sigma_i$, which is that the expected squared residual simply is the variance. We also drop the expectation around the first term, since as a function only of $\mu$ and $\mu_{i+1}$, it is no longer a function of $x_i$ which is the variable the expectation is under. We can then differentiate this expression with respect to $\sigma_i$ and solve for 0 to obtain the optimal variance.
\begin{align*}
    \frac{\partial \mathcal{F}}{\partial \sigma_i} &= \frac{\partial^2 \ln p(x_i | x{i+1})}{\partial x_i^2} - \sigma_i^{-1} \\
    & \frac{\partial \mathcal{F}}{\partial \sigma_i} = 0 \implies \sigma_i = {\frac{\partial^2 \ln p(x_i | x{i+1})}{\partial x_i^2}}^{-1} \numberthis
\end{align*}
Given this analytical result, there is no point optimizing $\mathcal{F}$ with respect to the variational variances $\sigma_i$, so our objective simply becomes,
\begin{align*}
    \mathcal{F} = \sum_{i=0}^L \ln p(\mu_i | \mu_{i+1}) \numberthis
\end{align*}
which is exactly the same result as obtained through the dirac delta approximation.

\section{Appendix B: Precision as Natural Gradients} 

One further fascinating connection between precisions and predictive coding comes through the fact that the precision parameters can be seen as implementing a natural gradient descent on the free-energy functional instead of simple gradient descent. Natural gradient methods \citep{amari1995information} augment the standard Euler-step gradient descent method with an additional matrix $\mathcal{G}$, which is the Fisher information with respect to the parameters of the descent. If we write the standard Euler-step gradient descent algorithm for some arbitrary parameters $\theta$ as,
\begin{align*}
    \theta_{t+1} = \theta_t + \eta \frac{\partial \mathcal{L}}{\partial \theta} \numberthis
\end{align*}
where $\mathcal{L}$ can be interpreted as either a loss function, or as the log-likelihood of a probabilistic model of the data given parameters. $\mathcal{L} = \ln f(o; \theta)$. For instance, the mean-squared error loss function can be interpreted as the log-likelihood of a Gaussian distribution, which is used in predictive coding. If this is the case, then the natural gradient descent rule would be
\begin{align*}
    \theta_{t+1} = \theta_t + \eta \mathcal{G}(\mathcal{L}, \theta) \frac{\partial \mathcal{L}}{\partial \theta} \numberthis
\end{align*}
where $\eta$ is a scalar learning rate and $\mathcal{G}({\mathcal{L}, \theta})$ is the Fisher information matrix of the parameters and likelihood function. The Fisher information is a central quantity in probability theory that can be intuitively interpreted as measuring the amount of information the data carries about the parameters $\theta$. It can be written in two equivalent ways -- as the variance of the score function where the score function is the gradient of the log likelihood, or equivalently as the negative expected Hessian of the likelihood function. 
\begin{align*}
        \mathcal{G}(\mathcal{L}, \theta) &= \mathbb{E}[\frac{\partial \ln L(o, \theta)}{\partial \theta}{\frac{\partial \ln L(o, \theta)}{\partial \theta}}^T] \\ 
        &= -\mathbb{E}[\frac{\partial^2}{\partial \theta^2}\ln \mathcal{L}(o,\theta)] \numberthis
\end{align*}

We will be principally concerned with the definition of the Fisher information in terms of the expected Hessian here. The intuition behind natural gradient descent is that `standard' gradient descent scales all elements of the gradient vector equally (implicitly $\mathcal{G}(\mathcal{L},\theta) = \mathbb{I}$). However, this is not necessarily the optimal way to move around in the loss landscape. Some directions may be changing much faster than other directions, and thus the effective step-size in some directions may be much larger or smaller in others. This could lead to inefficiently slow steps in regions where the loss landscape is changing slowly and thus the gradient vector is generally quite fixed, and to overly large, potentially destabilizing steps in regions where the loss landscape fluctuates rapidly. Natural gradients remedies this by modulating the gradient descent steps with information about the curvature (Hessian) of the loss function, taking smaller steps when the curvature is large (when the loss function is changing rapidly), and larger steps when the curvature is small (when the loss function is changing slowly). Theoretically this should improve the convergence properties of the descent, and improvements have been shown in a number of empirical studies \citep{amari1995information}, although explicit computation of the Hessian is challenging due to the size of the matrix as the number of parameters squared, which becomes intractable to store for extremely large models.

Importantly, in the case of predictive coding, the expected Hessian of the free-energy, with respect to the activity variables $\mu$ simply \emph{is} the precision, and thus, in effect, by optimizing the dynamics of $\mu$ using precision-weighted prediction errors, we are in effect performing a natural gradient descent. 
\begin{align*}
    \mathcal{G}(\mathcal{F}, \mu_l) &= \mathbb{E}[\frac{\partial^2}{\partial \mu_l^2}\mathcal{F}] \\ 
    &= \mathbb{E}[\frac{\partial^2}{\partial \mu_l^2} \big( (\mu_l - f(\theta_l \mu_{l+1}))^T \Sigma_l (\mu_l - f(\theta_l \mu_{l+1})) \big)] \\
    &= \mathbb{E}[\frac{\partial^2}{\partial \mu_l^2} \big( \mu_l \Sigma^{-1} \mu_l)] \\
    &= \mathbb{E}[\Sigma^{-1}_l] \\
    &= \Sigma^{-1} \numberthis
\end{align*}

Similarly, in the linear case, the Fisher information with respect to the parameters $\theta$ simply becomes the precision multiplied by the variance of the $\mu$s of the level above (recall the precision is effectively the variance of the prediction errors). $\epsilon$.
\begin{align*}
     \mathcal{G}(\mathcal{F}, \theta_l) &= \mathbb{E}[\frac{\partial^2}{\partial \theta_l^2}\mathcal{F}] \\ 
    &= \mathbb{E}[\frac{\partial^2}{\partial \theta_l^2} \big( (\mu_l - f(\theta_l \mu_{l+1})) \Sigma_l (\mu_l - f(\theta_l \mu_{l+1})) \big)] \\
    &= \mathbb{E}[\frac{\partial^2}{\partial \theta_l^2}  (\theta_l \mu_{l+1})^T \Sigma_l \theta_l \mu_{l+1}) ] \\
    &= \Sigma^{-1}_l \mathbb{E}[\mu_{l+1}\mu_{l+1}^T]
    \\ &= \Sigma^{-1}_l \mathbb{V}[\mu_{l+1}] \numberthis
\end{align*}
So that the Fisher information matrix for the parameters in the linear setting is the product of the variance of activities of the level above multiplied by the precision of the prediction errors at the level below. 
\section{Appendix C: Challenges for a Neural Implementation of Backpropagation by predictive Coding}

Despite the close links between predictive coding and backprop described in this review, there remain several challenges facing any direct predictive-coding based neural implementation of backprop. There are two primary challenges in directly translating such a scheme to neural circuitry. The first is the fixed prediction assumption, which requires the downward predictions to remain fixed throughout the optimization of the $\mu$s and the $\epsilon$s which is clearly unrealistic if such dynamical iterations are performed by the circulation of activity between cortical regions. It is potentially possible that there may exist neural circuitry that serves to store this activation during the iterative updates, however in general a more promising avenue may be to investigate the degree to which this assumption can be relaxed, and that the predictions can be allowed to change throughout iterations while maintaining learning performance. This is still an open question. The second challenge concerns the fact that the brain is fundamentally enmeshed in continuous time which the iterative nature of the convergence to backpropagation handles poorly. The world is not stationary while the brain undertakes its dynamical iterations -- this issue also affects standard predictive coding as well, since the sensory data is changing even while the prediction errors are being minimized. Intuitively, this would mean that fast convergence is required, and can often be achieved with fixed predictions, although much work remains to be assess the consequences of this assumption.

An additional interesting question is how to map the reverse-mode supervised learning paradigm directly to the brain, as it requires that the `loss function' be at the bottom of the hierarachy, and the recipient of the downward predictions. Taking a naive interpretation of sensory systems, then this would imply that the loss function is some sort of reconstruction or autoregressive loss on the sensory data, which is certainly plausible. Alternatively, we could consider the standard predictive coding model to be inverted \emph{within} the brain, which would imply that there is a feedforward sweep of upwards predictions, with feedback connectivity relaying downwards prediction errors, a view much closer to the standard `feedforward feature detector' view of sensory processing, where the loss would become some kind of either supervisory or contrastive loss in abstract high-level areas, or may be an intrinsically generated reward signal. This would require the inversion of the standard process theory \citep{bastos2012canonical}, such that superficial laminar layers carried feedforward predictions and the deep layers carried feedback prediction errors. Finally, both directions could be implemented simultaneously as a `hybrid' predictive coding model, which is setup in such a way that there is both top-down and bottom-up propagation, such that the dynamics of the error units attempt to reach a compromise between the bottom up and the top-down prediction errors. Such methods may be considered to be performing bottom-up and top-down backpropagation simultaneously, however in effect the top-down and bottom-up errors end up interfering with each other and thus convergence to backpropagated gradients is not achieved in either direction, but may be approximated sufficiently to enable effective learning.

A final note is also that while the current derivation only applies to backpropagation against a single fixed i.i.d sample, in practice brains must solve a much more challenging credit assignment problem through time, in which neural activity now can affect outcomes in the future, and therefore the credit must be backpropagated through time as well as space. The standard backpropagation algorithm can solve this through an extension called backpropagation through time (BPTT) which unrolls the computational graph across time and then backpropagates through the unrolled graph. However, this explicitly requires the direct propagation of information backwards in time, which is clearly implausible for neural circuits. How to achieve backpropagation through time, or approximations to it, is currently an open and exciting research question \citep{lillicrap2020backpropagation}, and there have been several promising approaches which propose online algorithms for computing the required quantities \citep{williams1989experimental,zenke2018superspike,bellec2020solution}. These require the notion of an eligibility-trace, which essentially accumulates information about the current state and supervisory signal in order to make correct credit assignments in the future.

\section{Appendix D: Kalman Filter Derivations}
Here we present full derivations for the weight updates for the predictive coding Kalman Filter.
\begin{align*}
    \frac{dL}{dA} &= \frac{d}{dA}[-2\mu_{t+1}^T \Sigma_1 A\mu_t + \mu_t^T A^T \Sigma_1 A\mu_t + \mu_t^T A^T \Sigma_1 Bu_t + u_t^TB^T \Sigma_1 A \mu_t] & \\
    &= -2\Sigma_1^T \mu_{t+1} \mu_t^T + \Sigma_1^T A\mu_t \mu_t^T \Sigma_1 A \mu_t\mu_t^T + \Sigma_1 Bu_t\mu_t^T + \Sigma_1^T Bu_t\mu_t^T \\
    &= -\Sigma_1[\mu_{t+1} - A\mu_t - Bu_t]\mu_t^T \\
    &= -\Sigma_1 \epsilon_x \mu_t^T \numberthis
\end{align*}
And, for the B (control) matrix.
\begin{align*}
    \frac{dL}{dB} &= \frac{dL}{dB}[2u_t^TB^T\Sigma_1 A\mu_t + u_t^TB^T\Sigma_1 Bu_t - 2\mu_{t+1}^T \Sigma_1 Bu_t] & \\
    &= (\Sigma_1 + \Sigma_1^T)Bu_t u_t^T + 2\Sigma_1 A \mu_t u_t^T - 2 \Sigma_1 \mu_{t+1} u_t^T \\
    &= - \Sigma_1[\mu_{t+1} - A\mu_t - Bu_t]u_t^T \\
    &= - \Sigma_1 \epsilon_x u_t^T \numberthis
\end{align*}
And, finally, for the C matrix, which is the observation matrix.
\begin{align*}
    \frac{dL}{dC} &= \frac{dL}{dC}[-2\mu_{t+1}^TC^T \Sigma_2 o_{t+1} + \mu_{t+1}^T C^T \Sigma_2 C \mu_{t+1}] &\\
    &= -2\Sigma_2 o_{t+1}\mu_{t+1}^T + 2\Sigma_2 C\mu_{t+1}\mu_{t+1}^T \\
    &= -\Sigma_2[o_{t+1} - C\mu_{t+1}]\mu_{t+1}^T \\
    &= -\Sigma_2 \epsilon_o \mu_{t+1}^T \numberthis
\end{align*}

\end{document}